\documentclass[runningheads]{llncs}

\usepackage{eccv}

\usepackage{eccvabbrv}

\usepackage{graphicx}
\usepackage{booktabs}
\usepackage{algorithm}
\usepackage{algpseudocode}
\usepackage{wrapfig}
\usepackage{tikz}
\usepackage{multirow}
\usepackage{bibunits}
\defaultbibliographystyle{splncs04}

\usepackage[accsupp]{axessibility}  

\usepackage{hyperref}

\usepackage{orcidlink}

\begin{document}
\begin{bibunit}
\title{Domain Adaptation with Adaptive Imagination for Visual Reinforcement Learning \\ under Limited Target Data} 

\titlerunning{AIDA}

\author{Hyunwoo Park\inst{1}\orcidlink{0009-0005-4825-9307} \and
Sang-Hyun Lee\inst{2}\orcidlink{0009-0007-3360-4758}}

\authorrunning{H.~Park and S.~Lee}

\institute{STRADVISION, Seoul, Republic of Korea \\
\email{hyunwoo.park@stradvision.com} \and
Department of Mobility Engineering, Ajou University, Suwon, Republic of Korea
\email{sanghyunlee@ajou.ac.kr}}

\maketitle

\begin{abstract}
Sim-to-real transfer remains a major obstacle for reinforcement learning (RL), especially for vision-based control where image observations exacerbate the state-distribution shift between simulation and the real world. Domain adaptation (DA) is a promising remedy for this challenge. Prior sim-to-real DA works have demonstrated encouraging results, yet these approaches typically assume substantially more target data, which is not available in practice. Indeed, their performance degrades significantly when the target data budget is reduced. To address this challenge, we propose AIDA (\textbf{A}daptive \textbf{I}magination for \textbf{D}omain \textbf{A}daptation), a domain adaptation framework for visual reinforcement learning that addresses sim-to-real transfer under scarce target data without requiring additional interaction with the target environment. Our key idea is adaptive imagination: generating reliable and semantic imagination rollouts to augment limited target data. Specifically, AIDA employs a distribution-shift-aware discriminator that truncates rollouts when imagined transitions drift into low-confidence regions, so that only reliable transitions contribute to the augmentation. On these reliable transitions, AIDA introduces a self-consistency loss that cycles through state → image observation → state, penalizing discrepancies between the original and reconstructed states. This provides additional adaptation signals beyond the scarce target data. Our experiments demonstrate that adaptive imagination effectively truncates unreliable rollouts. By enforcing a self-consistency loss on the resulting reliable transitions, AIDA learns semantically meaningful state representations and outperforms baselines across five MuJoCo tasks and two Gymnasium-Robotics tasks.
\end{abstract}

\section{Introduction}
\label{sec:intro}
Deep reinforcement learning (RL) has achieved remarkable success in simulation, enabling agents to master complex control tasks from raw sensory inputs~\cite{mnih2013playing, chen2021interpretable}. However, deploying these learned policies in the real world remains a fundamental challenge: discrepancies between simulation and reality—known as the sim-to-real gap—often cause dramatic performance degradation~\cite{zhao2020sim}. This gap can be broadly attributed to two factors: dynamics mismatch and state-distribution shift. In particular, for vision-based RL policies, state-distribution shift often plays a larger role. This is because pixel observations are high-dimensional and highly sensitive to changes in appearance and sensing conditions, which exacerbates the sim-to-real gap. In this work, we focus on mitigating this state-distribution shift for vision-based RL policies.

To cope with such distribution shifts, three families of methods are commonly used: domain randomization (DR), domain generalization (DG), and domain adaptation (DA). DR\cite{tobin2017domain, james2017transferring, pinto2017asymmetric, james2019sim} augments the simulator to train a single robust policy, hoping that the resulting variability covers factors of variation in the target environment. However, when the test domain is truly unknown or the shift is large, such coverage cannot be guaranteed. DG\cite{choi2023environment, wang2021unsupervised, hansen2021generalization} 
aims to learn policies that transfer to previously unseen test environments 
without target-domain supervision, typically by encouraging domain-agnostic 
representations and invariances (e.g., via attention/invariance objectives 
or augmentation-consistent regularization) rather than explicitly matching 
to a specific target domain. However, because the test domain is unknown 
by design, there is no guarantee that the training sources or augmentations 
capture the true factors of variation encountered at deployment; 
consequently, when the domain gap is large or includes unmodeled shifts, 
DG performance may degrade substantially—revealing a fundamental limitation 
of methods that must anticipate the target distribution without ever 
observing it \cite{wang2021unsupervised}.

DA\cite{hansen2020self, wang2024adapting, chen2021cross, sun2022transfer} is a promising alternative that directly leverages target-domain data to adapt to the specific target environment. This requires access to target data, but in return enables specialization to the particular deployment setting, even when the underlying causes of the shift are unknown at training time. Prior DA methods have indeed demonstrated superior performance over DR and DG~\cite{wang2024adapting, hansen2020self}. Nevertheless, they typically assume access to sufficient target data, and their performance degrades significantly when this budget is reduced~\cite{chen2021cross, niu2024xted}. Moreover, most prior DA methods adopt an image-to-image setting~\cite{hansen2020self, wang2024adapting, sun2022transfer}, which requires training RL agents in image-based simulation---a process that is usually harder~\cite{kaiser2019model, yarats2021improving} and can slow down RL by up to 20$\times$~\cite{xia2018gibson}. These two practical constraints remain underexplored despite their practical relevance. 

To address these challenges, we propose \textbf{AIDA} (\textbf{A}daptive \textbf{I}magination for \textbf{D}omain \textbf{A}daptation), a DA framework for visual RL that operates under scarce target data without additional target interaction. Our key idea is \emph{adaptive imagination}: augmenting limited target data with reliable imagination rollouts obtained via discriminator-gated truncation, and optimizing a self-consistency objective on these rollouts to learn semantically meaningful representations. Specifically, AIDA adopts a cross-modality setting (sim-state $\rightarrow$ real-image) that avoids the overhead of image-based RL training. A learned dynamics model produces policy-conditioned rollouts starting from target-inferred states, augmenting the limited target data with synthetic transitions. However, longer rollouts accumulate model error and drift from the target manifold. Therefore, a three-way discriminator assesses each imagined transition and truncates the rollout when confidence drops, so that only reliable, target-aligned transitions contribute to training. This adaptive truncation is critical, as imagination reliability varies across both tasks and individual states, making any fixed imagination horizon suboptimal. On the resulting reliable transitions, we introduce a \emph{self-consistency loss} that enforces a state $\rightarrow$ image $\rightarrow$ state cycle on discriminator-gated imagination rollouts. Unlike prior cycle objectives limited to short, data-supported predictions, our self-consistency loss extends cycle supervision to multi-step, \emph{policy-conditioned} imagination rollouts, extracting additional training signal specifically from states the agent is likely to visit and thereby promoting semantically meaningful state representations.

We evaluate AIDA across five MuJoCo tasks and two Gymnasium-Robotics tasks with scarce pre-collected target data, where it consistently outperforms baselines, demonstrating its effectiveness in practically important yet challenging regime. Our contributions are:
\begin{itemize}
    \item[(i)] \textbf{AIDA} A novel paradigm that augments limited target data with reliable synthetic transitions and leverages self-consistency to learn semantically meaningful representations. 
    \item[(ii)] \textbf{Discriminator-gated imagination.} A mechanism for generating only reliable imagined transitions by dynamically truncating imagination rollouts based on the discriminator's domain classification signal.
    \item[(iii)] \textbf{Self-consistency via imagination-augmented data.} A scheme that learns semantically meaningful representations by enforcing a self-consistency loss on synthetic policy-conditioned imagination rollouts, providing additional supervision under limited target data.
    \item[(iv)] \textbf{Experimental evaluation.} Comprehensive experiments across five MuJoCo tasks and two Gymnasium-Robotics tasks demonstrate that imagined transitions progressively become unreliable as the rollout horizon grows and that discriminator-gated truncation yields higher returns than any fixed-horizon alternative, enabling AIDA to consistently outperform baseline DA methods in the low-data regime.
\end{itemize}

\section{Related Work}

Bridging the sim-to-real gap has been a longstanding challenge in RL, and a number of approaches have been proposed to address the shift in state distributions between source and target domains. Broadly, these methods fall into three categories: domain randomization (DR), domain generalization (DG), and domain adaptation (DA).

\textbf{Domain Randomization (DR).}
DR trains an agent in simulation under a wide range of environment parameters—such as textures, lighting, camera poses, and object appearances—with the goal that such visual variations will cover the target-domain state distribution. In vision-based settings, this typically involves randomizing camera poses, lighting conditions, object placements, textures (including non-realistic ones), and backgrounds~\cite{tobin2017domain, james2017transferring, pinto2017asymmetric, james2019sim}. The resulting randomized images can be fed directly to the agent for policy learning, or first passed through an image-translation generator to produce more realistic observations before being used for training.

However, DR often requires generating a large number of variants to encourage the policy to learn invariant representations, which makes training computationally expensive. More importantly, the randomization space (i.e., which factors to vary and their ranges) is manually specified by the practitioner; therefore, target domains may contain additional, unforeseen factors that are not included in the chosen variants. As a result, the assumed variants may fail to fully cover the target-domain distribution, leading to degraded transfer performance~\cite{hansen2020self}.

\textbf{Domain Generalization (DG).} While DR broadens the training distribution by randomizing simulator parameters to cover a particular  target domain, DG assumes no target-domain data at training time and seeks to generalize to unseen domains by learning domain-invariant representations from diverse sources/augmentations. In vision-based RL, DG is commonly implemented by shaping the visual representation to be insensitive to appearance changes—for example, through visual augmentations~\cite{hansen2021generalization}, representation-level consistency and foreground extraction~\cite{wang2021unsupervised}, or feature factorization/reconstruction objectives~\cite{choi2023environment}. However, because the test domain is unknown, there is no guarantee that the training sources/augmentations capture the true factors of variation encountered at deployment; consequently, when the domain gap is large or includes unmodeled shifts, DG performance may degrade substantially.

\textbf{Domain Adaptation (DA).} Unlike DR and DG, DA assumes access to a target environment and adapts a source-trained policy to the target domain before deployment or at test time~\cite{hansen2020self, wang2024adapting, chen2021cross, sun2022transfer}. Because the adaptation is driven by target-domain data, DA can reduce the source--target shift even when the primary factors behind the domain gap are unknown, focusing updates on discrepancies that are actually observed in the target domain. 

Concretely, a common design is to adapt \emph{what the policy sees} (i.e., the observation/feature representation) so that source and target become consistent, or to adapt the \emph{policy itself} using target-domain experience. For example, PAD~\cite{hansen2020self} performs test-time adaptation by updating the visual encoder with an inverse-dynamics self-supervised objective, while CODAS~\cite{chen2021cross} uses a GAN-based mapping to align target visual observations with source-side privileged information. PRFT~\cite{wang2024adapting} instead adapts the policy during deployment by fine-tuning with predicted rewards computed in the target domain, enabling reward-free adaptation. Sun et al. \cite{sun2022transfer} transfer source-trained latent transition and reward models as fixed regularizers to guide representation learning in the target domain, effectively encouraging target features to be compatible with source dynamics priors. 
Despite their effectiveness, these methods typically assume access to substantial target data, which is often impractical in real-world deployment where data collection is costly and limited. In contrast, our work explicitly targets the scarce-data regime, which remains underexplored despite its practical importance.
\section{Preliminaries}
\label{sec:preliminaries}

\subsection{Markov Decision Process.}
We consider a Markov Decision Process defined by a tuple $(\mathcal{S}, \mathcal{A}, p, r, \gamma)$, where $\mathcal{S}$ is the state space, $\mathcal{A}$ is the action space, $p(s' \mid s, a)$ is the transition dynamics, $r(s, a)$ is the reward function, and $\gamma \in (0,1)$ is the discount factor. The goal of RL is to learn a policy $\pi(a \mid s)$ that maximizes the expected cumulative return $\mathbb{E}\left[\sum_{t=0}^{T} \gamma^t r(s_t, a_t)\right]$. In visual RL, the agent does not observe the underlying state $s$ directly; instead, it receives a high-dimensional image observation $o \in \mathcal{O}$, and must learn to act from pixels.

A dynamics model $f_\omega: \mathcal{S} \times \mathcal{A} \rightarrow \mathcal{S}$ is a parameterized approximation of the transition dynamics $p(s' \mid s, a)$. Given a state $s_t$ and action $a_t$, it predicts the next state $\tilde{s}_{t+1} = f_\omega(s_t, a_t)$. We learn $\omega$ by minimizing a one-step prediction loss such as $\mathcal{L}_{wm} = \mathbb{E}_{(s,a,s')\sim p(s'\mid s,a)}
\!\left[ \left \| f_{\omega}(s,a) - s' \right\|_2^2 \right].$ By repeatedly applying $f_\omega$ under a policy $\pi$, the agent generates multi-step rollouts $\tilde{s}_{t+1}, \tilde{s}_{t+2}, \ldots$ entirely in state space without interacting with the environment---a process referred to as \emph{imagination}~\cite{hafner2019dream, sutton1991dyna}. Since the dynamics model is an approximation, prediction error accumulates over successive steps, making longer rollouts less reliable.

\subsection{Domain Adaptation in RL as Variational Inference.}
\label{pre:gan}
When the source domain provides low-dimensional states while the target domain provides image observations, a mapping function must project target observations into the source state space to enable policy transfer. We build on the trajectory-level variational inference framework of~\cite{chen2021cross} for aligning inferred and source state distributions. This framework models a \emph{generation process} via an auto-regressive observation model $p_\theta(o_t \mid \hat{s}_t, o_{t-1})$ that renders target observations from inferred states $\hat{s}_t$ and the previous observation $o_{t-1}$, and an \emph{inference process} via a mapping function $q_\phi(s_t \mid \hat{s}_{t-1}, a_{t-1}, o_t)$ that infers a mapped state $\hat{s}_t$ by combining the previous inferred state 
$\hat{s}_{t-1}$, the previous action $a_{t-1}$, and the target observation $o_t$. Applying step-wise inference over a full trajectory defines a variational posterior $q_\phi(\tau^s \mid \tau^o)$ over the source state--action trajectory $\tau^s = \{(s_1,a_1),\ldots,(s_T,a_T)\}$ given the target observation--action trajectory $\tau^o = \{(o_1,a_1),\ldots,(o_T,a_T)\}$. The resulting ELBO is:
\begin{equation*}
\max_{\phi,\theta}\; \mathbb{E}_{\tau^o}\!\Big[
\mathbb{E}_{\hat{\tau}^s \sim q_\phi(\tau^s \mid \tau^o)}\big[\log p_\theta(\tau^o \mid \hat{\tau}^s)\big] - D_{\mathrm{KL}}\!\left(q_\phi(\tau^s \mid \tau^o)\;\|\; p(\tau^s)\right) \Big],
\end{equation*}
where $\hat{\tau}^s$ denotes an inferred state--action trajectory sampled from $q_\phi$, the first term is a reconstruction loss $\mathcal{L}_\mathrm{recon}$ enforcing visual consistency, and the second term aligns the inferred trajectory distribution with the source prior. Since the KL term is intractable, it is approximated with the adversarial loss~\cite{nowozin2016f}:
a binary discriminator $D_\eta$ distinguishes source trajectories from inferred ones, while $q_\phi$ acts as the generator. We denote this loss as $\mathcal{L}_{\mathrm{adv}}$:
\begin{equation}
\begin{aligned}
\min_{\phi}\max_{\eta}\;
&\mathbb{E}_{\tau^s \sim \mathcal{D}_{src}}\!\left[\log D_\eta(\tau^s)\right]
+ \mathbb{E}_{\tau^o \sim \mathcal{D}_{tgt},\,\hat{\tau}^s \sim q_\phi(\cdot\mid\tau^o)}
\!\left[\log\big(1-D_\eta(\hat{\tau}^s)\big)\right].
\end{aligned}
\label{eq:gan_align}
\end{equation}
Together with the reconstruction loss, this defines the trajectory alignment loss $\mathcal{L}_\mathrm{align} = \mathcal{L}_\mathrm{recon} + \alpha\mathcal{L}_\mathrm{adv}$, where $\alpha$ controls the relative weight of the adversarial alignment term. Trajectory-level alignment is essential because state-level matching alone is ill-posed: a target observation may be mapped to an incorrect but plausible source state that is indistinguishable without sequential context. 
\begin{figure}[tb] 
\centering 
\includegraphics[width=0.92\linewidth]{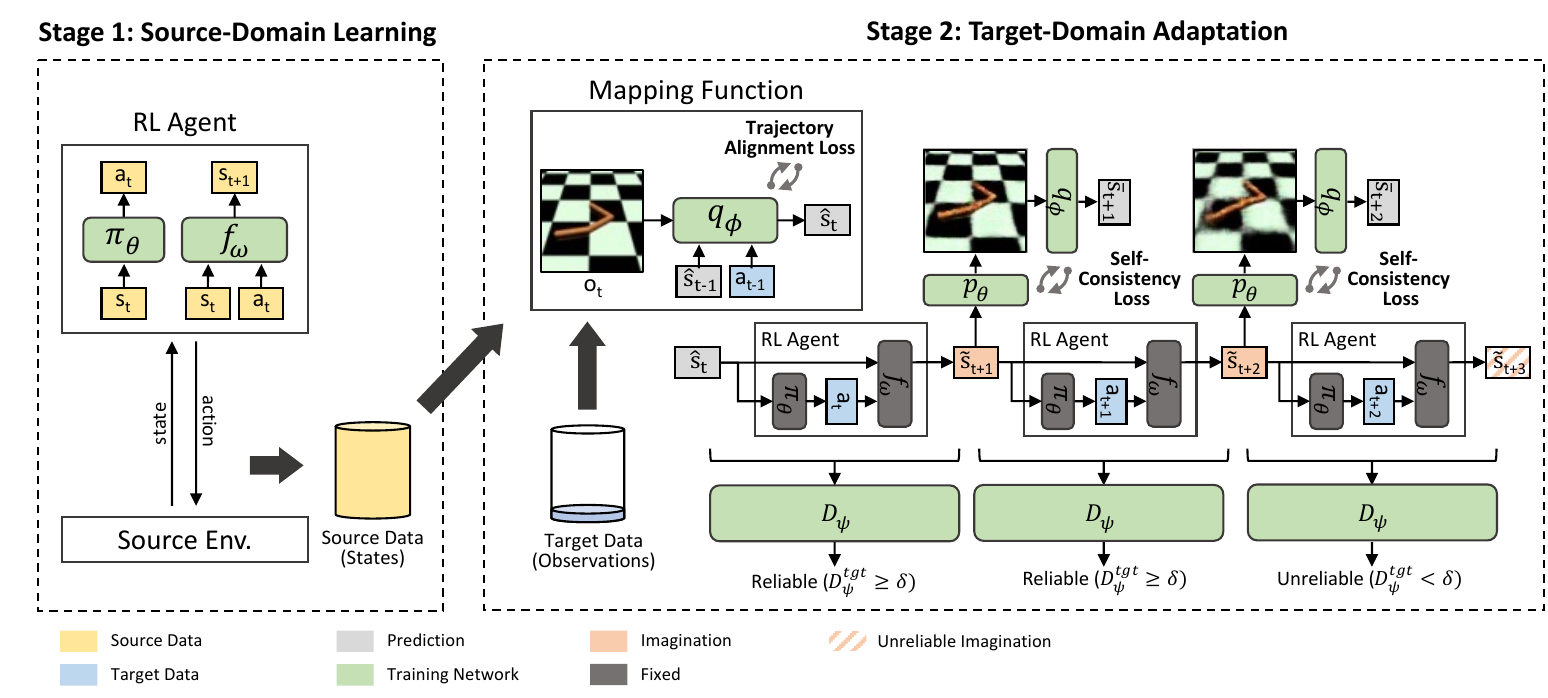} 
\caption{\textbf{Overview of AIDA.} The framework proceeds in two stages.
\textbf{Stage 1 (source-domain learning):} the agent interacts with the source environment to train a dynamics/world model $f_{\omega}$ and a policy $\pi$ via dynamics learning and reinforcement learning.
\textbf{Stage 2 (target-domain adaptation):} given scarce target image trajectories $\mathcal{D}_{\mathrm{tgt}}$, we freeze $f_{\omega}$ and $\pi$ and generate \emph{reliable, adaptively truncated} imagined trajectories $\mathcal{D}_{\mathrm{img}}$.
A mapping function $q_{\phi}$ and an observation model $p_{\theta}$ are then optimized with a trajectory alignment loss $\mathcal{L}_{\mathrm{align}}$ and a self-consistency loss $\mathcal{L}_{\mathrm{sc}}$ (state $\rightarrow$ image $\rightarrow$ state) to learn target-aligned imagination under limited target data.}
\label{fig:overview} 
\end{figure}
\section{Method}
\label{sec:method}

AIDA operates in two stages. In the first stage, a state-based policy $\pi$ and a dynamics model $f_\omega$ are trained in the source domain with full environment access. In the second stage, given only a small pre-collected dataset of target-domain image trajectories $\mathcal{D}_{tgt}$, AIDA adapts the source-trained policy to the target domain without any additional target interaction. The adaptation builds on the trajectory alignment framework described in \cref{sec:preliminaries}, optimizing the trajectory alignment loss. However, the core challenge is that scarce target data provides insufficient coverage for alignment. AIDA addresses this through adaptive imagination: a learned dynamics model generates synthetic rollouts from target-inferred states, a three-way discriminator truncates these rollouts when they drift from the target manifold, and a self-consistency loss on the resulting reliable transitions provides additional supervision for learning semantically meaningful representations. Our overall workflow is summarized in \cref{fig:overview}. We begin by formally defining the problem setting (\cref{sec:problem_formulation}), then describe each component in detail. Additional implementation details are provided in the supplementary material.
\subsection{Problem Formulation}
\label{sec:problem_formulation}
We consider domain adaptation for visual RL between a source and a target domain. We distinguish the underlying \emph{state} $s \in \mathcal{S}$, which specifies the physical configuration of the system, from the \emph{observation} received by the agent. Following common sim-to-real DA formulations~\cite{chen2021cross, hansen2020self, sun2022transfer}, we assume that the source and target share the same state space $\mathcal{S}$, action space $\mathcal{A}$, transition dynamics, and reward function, but differ in their observation spaces.

A straightforward approach to sim-to-real visual adaptation is \emph{image-to-image adaptation}, which encodes source and target observations into a shared latent feature space and aligns the resulting representations so that a policy trained on the aligned features can transfer across domains~\cite{hansen2020self, wang2024adapting, sun2022transfer}. However, visual-based RL is notorious for unstable training~\cite{kaiser2019model, yarats2021improving} and expensive computation~\cite{xia2018gibson}, making image-to-image approaches impractical when interaction and computation budgets are limited. Moreover, most existing domain adaptation methods assume access to a large amount of target-domain data~\cite{chen2021cross}, which is often unrealistic since collecting real-world data is far more costly than generating simulated data.

In this work, we consider a challenging setting that addresses both of these limitations. First, we adopt a \emph{cross-modality} problem setting. The policy is trained in the source domain using low-dimensional simulator states $s \in \mathcal{S}$, while the target domain provides only high-dimensional image observations $o \in \mathcal{O}$. Simulators typically provide privileged access to the full state, which enables substantially more efficient policy learning. However, this design turns deployment into a cross-modality transfer problem ($\text{sim(state)} \rightarrow \text{real(image)}$) and enlarges the domain gap compared to image-to-image adaptation, since the policy must operate from a different sensing modality at test time. Second, we consider a \emph{scarce target data} regime. Only a small number of target-domain trajectories are available, and no additional target interaction is permitted during adaptation. This makes the adaptation problem substantially harder, as the alignment module must generalize from minimal target samples.

Consequently, domain adaptation in our setting reduces to learning an image-to-state mapping $q_\phi \colon \mathcal{O} \rightarrow \mathcal{S}$ from scarce target data, projecting target observations into the source state space and enabling direct deployment of the source policy in the target domain as $\pi(q_\phi(o))$. The target dataset $\mathcal{D}_{tgt}$ for training $q_\phi$ consists of trajectories of image observations paired with executed actions, \eg, tuples $(o_t, a_t, o_{t+1})$, and contains no privileged target states.


\subsection{Discriminator-Gated Imagination}
When target data is scarce, the alignment objective in \cref{eq:gan_align} alone provides insufficient supervision. To obtain additional training signal, we leverage the learned dynamics model to generate imagination rollouts from target-inferred states, synthesizing auxiliary transitions beyond the limited target dataset. A key question, however, is how long to roll out. If the rollout horizon is too long, imagined transitions drift into regions unsupported by target data, where the mapping is unreliable. Additionally, the dynamics model accumulates prediction error over successive steps, further degrading rollout quality. If it is too short, too few synthetic transitions are generated to supplement the scarce target data. Therefore, effective adaptation requires choosing an imagination horizon that is long enough to provide useful supervision while remaining within regions where the imagined transitions are reliable.

To address this, we introduce a three-way discriminator $D_\psi$ (distinct from the binary alignment discriminator $D_\eta$ in~\cref{pre:gan}) trained to classify transitions as (i)~\emph{target-inferred}, (ii)~\emph{source}, or (iii)~\emph{imagined}. Since the discriminator is trained on real target transitions, it assigns high target-likeness scores to imagined transitions that resemble the target data and low scores to those that have drifted away, making its output a natural indicator of rollout reliability. We use this score as an online confidence signal to adaptively truncate rollouts, continuing while the score remains high and stopping when it drops, so that only reliable transitions contribute to training. The three-way formulation is important because a binary \emph{source-vs-target} discriminator scores transitions that drift outside the support of \emph{both} domains ambiguously (often near $0.5$), which may be mistakenly interpreted as ``moderately target-like'' even though they are unlike either distribution. Introducing an explicit \emph{imagined} class allows the discriminator to recognize such out-of-support rollouts as \emph{imagined/OOD} rather than assigning a misleading intermediate probability.

Utilizing the three-way discriminator as an online confidence signal, we now describe how imagination rollouts are generated and adaptively truncated during training. Given a target-initialized state $\hat{s}_t \sim q_\phi({s}_t \mid \hat{s}_{t-1}, a_{t-1}, o_t)$, we roll out imagined states $\tilde{s}_{t+1},\tilde{s}_{t+2},\ldots$ using the dynamics model under the frozen policy $\pi$.
Let $D_\psi^{\mathrm{tgt}}(x)\triangleq P_\psi(y=\mathrm{target\text{-}inferred}\mid x)$ denote the discriminator's predicted probability of the \emph{target-inferred} class.
At each step, we use $D_\psi^{\mathrm{tgt}}(\cdot)$ as a \emph{target-likeness} score and define the adaptive rollout horizon as the first step at which this score falls below a threshold $\delta \in (0, 1)$, limited to a maximum horizon $K_{\max}$,
\begin{equation}
K^* = \min \left( \min \left\{ k \geq 0 \;\middle|\; 
D_\psi^{\mathrm{tgt}}\!\left(\tilde{s}_{t+k},\, a_{t+k},\, 
\tilde{s}_{t+k+1}\right) < \delta \right\},\, K_{\max} \right),
\end{equation}
so that imagination proceeds for steps $k = 0, 1, \ldots, K^*-1$.
A higher $D_\psi^{\mathrm{tgt}}$ indicates that the rollout remains within the target-aligned manifold, whereas a low score signals drift into a low-confidence region where self-consistency training would be counterproductive.

Discriminator-gated imagination offers more than simply finding an appropriate rollout length, because the truncation decision is made per-state rather than globally. This is crucial since each state from which imagination begins leads to a different rollout trajectory, and the maximum length of reliable imagination varies accordingly. In practice, we observe that the rollout horizon varies significantly depending on the state (see~\hyperref[drift]{5.2.Q2}), confirming that a single fixed $K$ cannot capture this per-state variability, and that the adaptive scheme consistently outperforms all fixed-horizon alternatives (see~\hyperref[fixed_comparison]{5.2.Q3}).


\subsection{Self-Consistency via Imagination-Augmented Data}
Here we describe how to fully leverage the reliable imagined transitions obtained from discriminator-gated truncation. Our key idea is to impose a self-consistency constraint on these imagined transitions, requiring that a state decoded into an image and re-inferred should recover the original. This cycle-consistency idea has been explored in latent space~\cite{huang2018multimodal,zhu2017toward}, but only on real data points. Our formulation applies this constraint to \emph{imagined} states from multi-step rollouts, enabling supervision even in regions where no real target data exists. For each reliable imagined state $\tilde{s}_{t+k}$ ($k \leq K^*$) admitted by the discriminator gate, we generate an imagined observation $\tilde{o}_{t+k} \sim p_\theta(\cdot \mid \tilde{s}_{t+k}, \tilde{o}_{t+k-1})$ and re-infer a state $\bar{s}_{t+k} \sim q_\phi({s}_{t+k} \mid \tilde{s}_{t+k-1}, a_{t+k-1}, \tilde{o}_{t+k})$. We penalize discrepancies between the re-inferred and original imagined states,
\begin{equation}
\mathcal{L}_{\mathrm{sc}}
=
\frac{1}{K^*}
\sum_{k=1}^{K^*}
\left\|
\bar{s}_{t+k} - \tilde{s}_{t+k}
\right\|_2^2 .
\end{equation}

This loss encourages the encoder to capture semantically meaningful state information that is \emph{visually grounded}. Factors that cannot be rendered into target-like observations and re-inferred are suppressed, while state-relevant factors that remain consistent under the decode$\rightarrow$re-infer cycle are reinforced. Furthermore, since our consistency constraint is enforced over \emph{multi-step} imagination rather than short, data-supported predictions, it yields extra supervision across successive rollout steps. The \emph{policy-conditioned} rollouts further concentrate this supervision on \emph{reachable} states the current policy is likely to encounter. Because reliable imagination can generate many such multi-step state--observation pairs without additional target interaction, the self-consistency objective provides dense auxiliary supervision beyond scarce real target data, stabilizing representation alignment in regions where real samples are insufficient.

Combining the trajectory alignment objective from \cref{sec:preliminaries} with the self-consistency loss, the overall training objective for Stage~2 is:
$
\mathcal{L} = \mathcal{L}_{align} + \lambda \, \mathcal{L}_{sc},$
where $\lambda$ controls the relative weight of the self-consistency regularization.
\section{Experiments}
\label{sec:experiments}

We design our experiments to investigate the following questions: (1) Does AIDA outperform existing domain adaptation baselines under scarce target data? (2) Does the three-way discriminator reliably detect unreliable drifted transitions? (3) Does the adaptive imagination outperform fixed-horizon imagination, where unreliable imagined transitions are inevitably included? (4) Does the self-consistency loss encourage the encoder to learn state representations that faithfully capture the true physical configuration? We first describe the experimental setup, then address each question in \cref{performance_baseline}.

\subsection{Experimental Setup}
We evaluate AIDA on seven continuous control tasks: five tasks from the MuJoCo benchmark suite~\cite{todorov2012mujoco}, HalfCheetah, Hopper, Swimmer, Walker2d, and InvertedPendulum, and two tasks from Gymnasium-Robotics, Shadow Dexterous Hand Reach and Fetch Reach. Following the problem setting in~\cref{sec:problem_formulation}, the source domain provides low-dimensional states while the target domain provides only high-dimensional image observations, creating a large cross-modality gap.
The target data budget is severely restricted to reflect the practical reality that real-world data collection is costly and limited: only \textbf{50 trajectories} are available for all tasks---$1/6$ of the target data used in previous work~\cite{chen2021cross}. For InvertedPendulum, a relatively simple task, we reduce the trajectory length $T$ and image resolution to avoid performance saturation across all methods. The target trajectories are collected by executing the expert policy in the target environment, and no additional interaction with the target environment is permitted during adaptation, except for PAD, the online adaptation baseline. The source-domain policy $\pi$ is trained using Soft Actor-Critic (SAC)~\cite{haarnoja2018soft} with low-dimensional state inputs until convergence, then frozen throughout the adaptation procedure. 
Additional details are provided in the supplementary material.

We compare AIDA against five methods that span different approaches to bridging the observation mismatch:
\begin{itemize}
    \item \textbf{CODAS}~\cite{chen2021cross}: A GAN-based domain adaptation method that aligns target visual observations with source-side privileged state information via a variational inference framework with adversarial training. As the closest prior work to our setting, comparing against CODAS directly measures the benefit of adaptive imagination under the same problem setting.
    \item \textbf{GAN\_STACK}~\cite{chen2021cross}: A GAN-based domain adaptation method that stacks consecutive observations and feeds them to the alignment module, providing temporal context. This baseline tests whether simply providing more temporal information can substitute for the imagination-based augmentation that AIDA employs.
    \item \textbf{PAD}~\cite{hansen2020self}: A self-supervised test-time adaptation method originally proposed for adapting vision-based RL policies to unseen visual changes. Unlike the other offline baselines, PAD continuously interacts with the target environment during deployment. This baseline evaluates whether self-supervised online adaptation is sufficient under our larger state-to-image domain gap, compared to AIDA.

    \item \textbf{Behavior Cloning (BC)}: A supervised baseline that directly learns a mapping from target image observations to source states using paired data, without any adversarial or self-consistency objective. This serves as a reference for how far pure supervised learning can go with scarce paired data.
    \item \textbf{Oracle (Image)}: SAC combined with a convolutional autoencoder~\cite{yarats2021improving} trained jointly with the RL objective on target-domain image observations with \emph{unlimited} online interaction. This serves as an approximate upper bound for visual RL without cross-modal transfer, providing a reference point for the offline adaptation methods.
\end{itemize}

\subsubsection{Evaluation metrics.}
We report two complementary metrics:
\begin{itemize}
    \item \textbf{Root Mean Squared Error (RMSE)}: Measures the accuracy of the learned mapping $q_\phi$ by computing the root-mean-squared distance between the inferred states and the ground-truth states.
    \item \textbf{Return Ratio}: Measures the task performance of the adapted policy relative to the source-trained expert. Defined as $r_{\mathrm{ratio}} = r / r^*$, where $r$ is the cumulative return of the adapted policy deployed in the target environment, and $r^*$ is the average return of the source expert. A ratio of $1.0$ indicates full recovery of the source expert's performance.
\end{itemize}
\subsection{Experimental Results and Analysis}
\subsubsection{(Q1) Does AIDA outperform existing domain adaptation baselines under scarce target data?}
\label{performance_baseline}

\begin{table}[tb]
\centering
\caption{Comparison of domain adaptation methods under scarce target 
data. RMSE ($\downarrow$) and Return Ratio ($\uparrow$) report 
converged performance. A ratio of $1.0$ corresponds to the source 
expert's average return. Best results are in \textbf{bold}. 
Mean$\pm$std over 3 seeds.}
\label{tab:main_results}
\renewcommand{\arraystretch}{1.08}
\setlength{\tabcolsep}{4.5pt}

\resizebox{\linewidth}{!}{%
\begin{tabular}{@{}lccccccc@{}}
\toprule
\multicolumn{8}{@{}c@{}}{\textbf{(a) Return Ratio} ($\uparrow$)} \\
\midrule
Method 
& H.Cheetah & Hopper & Swimmer & Walker2d 
& Inv.Pend. & Shadow & Fetch \\
\midrule
Oracle (image)
& $1.638{\pm}.343$
& $0.591{\pm}.232$
& $0.304{\pm}.060$
& $0.196{\pm}.073$
& $0.989{\pm}.072$
& $0.991{\pm}.045$
& $0.975{\pm}.076$ \\

\midrule
BC
& $0.462{\pm}.011$
& $0.121{\pm}.018$
& $0.403{\pm}.042$
& $\mathbf{0.101{\pm}.013}$
& $0.319{\pm}.052$
& $0.836{\pm}.033$
& $0.453{\pm}.009$ \\

GAN\_STACK
& $0.340{\pm}.087$
& $0.056{\pm}.008$
& $0.261{\pm}.066$
& $0.026{\pm}.004$
& $0.063{\pm}.004$
& $0.645{\pm}.122$
& $0.823{\pm}.053$ \\

PAD
& $0.573{\pm}.075$
& $0.142{\pm}.023$
& $0.347{\pm}.058$
& $0.036{\pm}.005$
& $0.155{\pm}.012$
& $0.395{\pm}.088$
& $0.821{\pm}.044$ \\

CODAS
& $0.711{\pm}.098$
& $0.356{\pm}.086$
& $0.468{\pm}.069$
& $0.038{\pm}.009$
& $0.440{\pm}.038$
& $0.893{\pm}.059$
& $0.897{\pm}.007$ \\

\midrule
AIDA (ours)
& $\mathbf{0.810{\pm}.113}$
& $\mathbf{0.440{\pm}.142}$
& $\mathbf{0.512{\pm}.047}$
& $0.058{\pm}.018$
& $\mathbf{0.561{\pm}.049}$
& $\mathbf{0.931{\pm}.027}$
& $\mathbf{0.984{\pm}.012}$ \\
\bottomrule
\end{tabular}%
}

\vspace{0.75em}

\resizebox{\linewidth}{!}{%
\begin{tabular}{@{}lccccccc@{}}
\toprule
\multicolumn{8}{@{}c@{}}{\textbf{(b) RMSE} ($\downarrow$)} \\
\midrule
Method 
& H.Cheetah & Hopper & Swimmer & Walker2d 
& Inv.Pend. & Shadow & Fetch \\
\midrule
GAN\_STACK
& $3.006{\pm}.153$
& $0.862{\pm}.059$
& $1.360{\pm}.272$
& $2.525{\pm}.317$
& $0.328{\pm}.042$
& $0.452{\pm}.029$
& $0.025{\pm}.009$ \\

CODAS
& $1.550{\pm}.081$
& $0.540{\pm}.021$
& $0.482{\pm}.057$
& $2.180{\pm}.405$
& $0.108{\pm}.007$
& $0.425{\pm}.037$
& $0.017{\pm}.004$ \\

\midrule
AIDA (ours)
& $\mathbf{1.403{\pm}.116}$
& $\mathbf{0.523{\pm}.023}$
& $\mathbf{0.356{\pm}.015}$
& $\mathbf{1.784{\pm}.320}$
& $\mathbf{0.075{\pm}.004}$
& $\mathbf{0.391{\pm}.016}$
& $\mathbf{0.015{\pm}.008}$ \\
\bottomrule
\end{tabular}%
}

\end{table}

\begin{figure*}[tb]
  \centering
  \captionsetup[subfigure]{font=footnotesize}
  \captionsetup{skip=2pt}

  \begin{subfigure}[t]{0.32\textwidth}
    \centering
    \includegraphics[width=\linewidth]{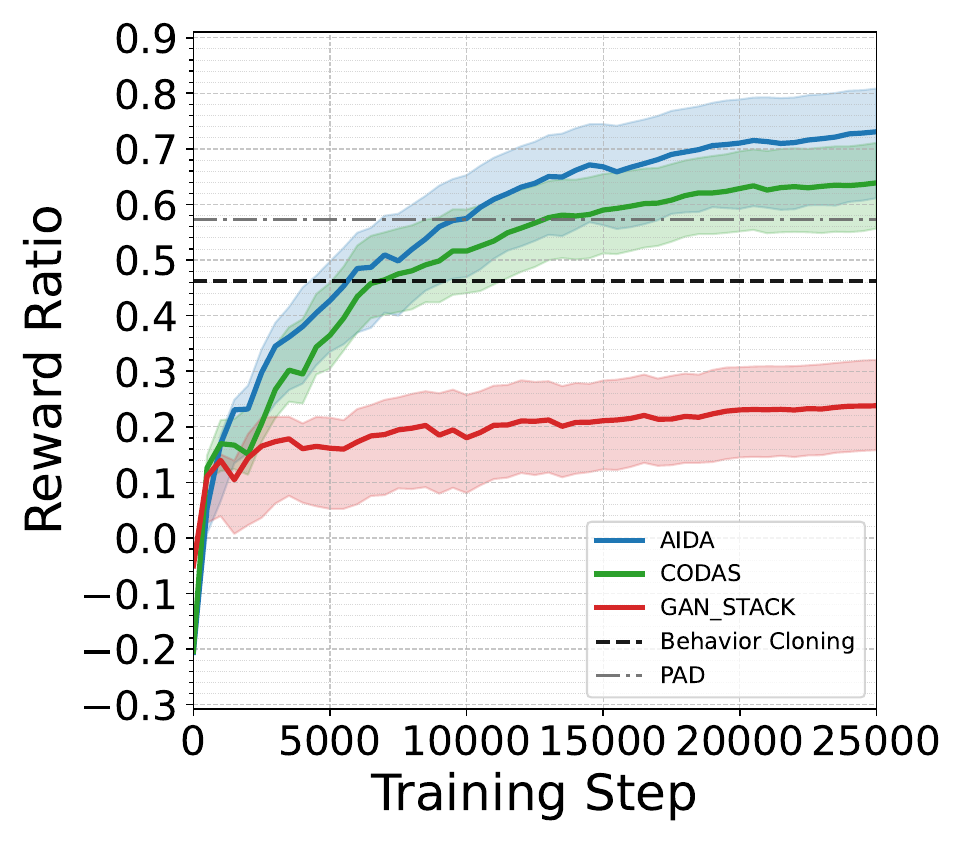}
    \caption{Return -- HalfCheetah}
    \label{fig:eval_halfcheetah}
  \end{subfigure}\hspace{0.006\textwidth}%
  \begin{subfigure}[t]{0.32\textwidth}
    \centering
    \includegraphics[width=\linewidth]{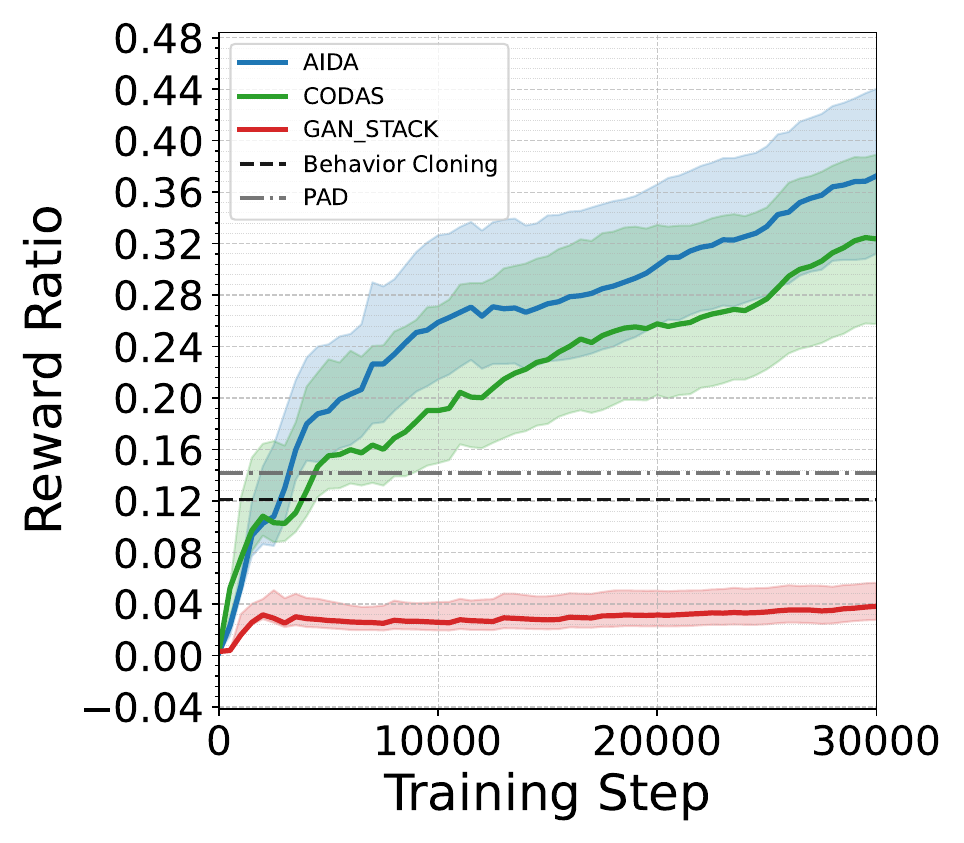}
    \caption{Return -- Hopper}
    \label{fig:eval_hopper}
  \end{subfigure}\hspace{0.006\textwidth}%
  \begin{subfigure}[t]{0.32\textwidth}
    \centering
    \includegraphics[width=\linewidth]{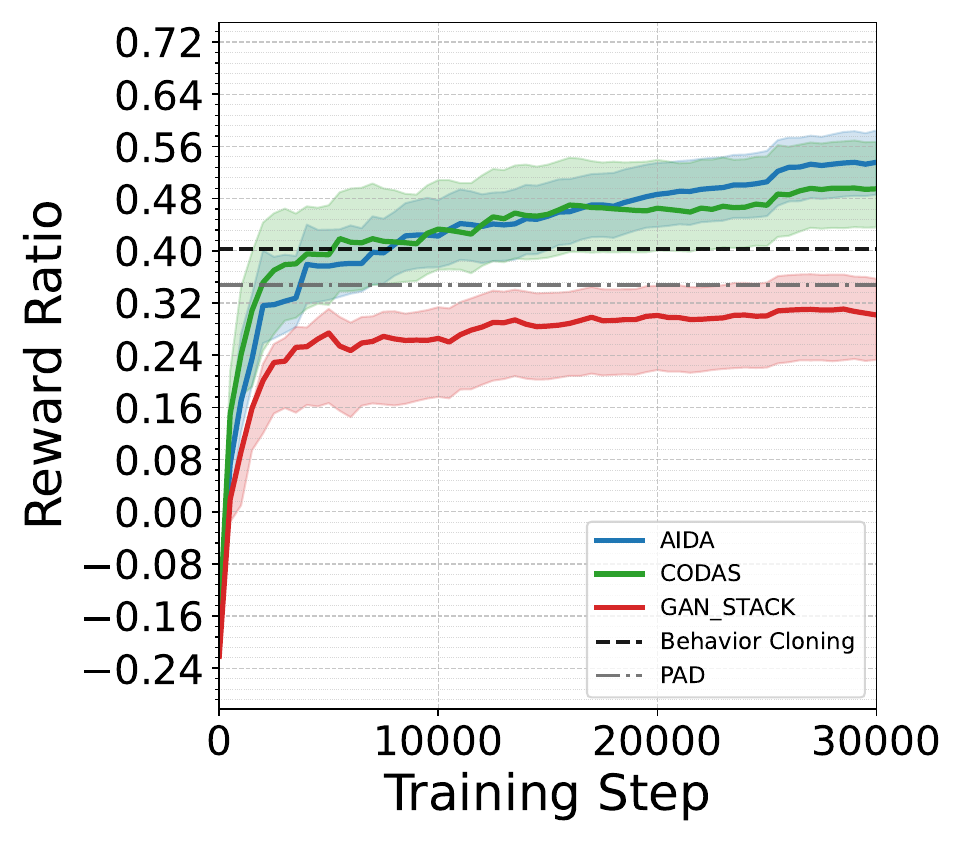}
    \caption{Return -- Swimmer}
    \label{fig:eval_swimmer}
  \end{subfigure}

  \vspace{0.2em}

  \begin{subfigure}[t]{0.32\textwidth}
    \centering
    \includegraphics[width=\linewidth]{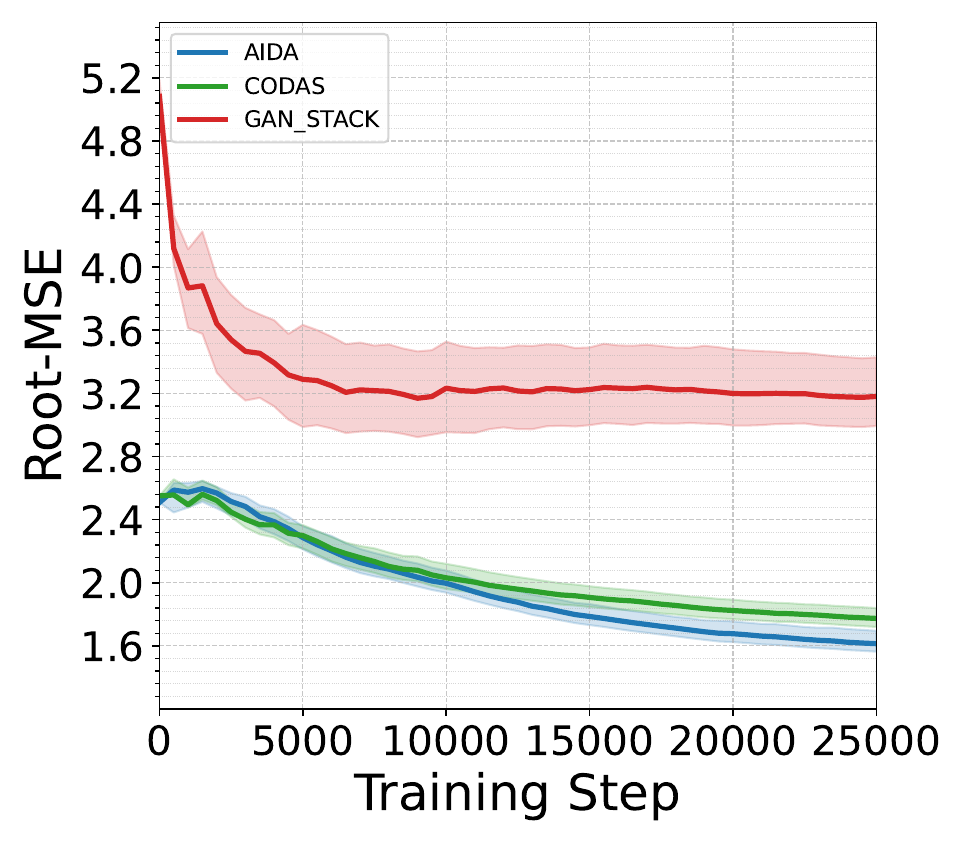}
    \caption{RMSE -- HalfCheetah}
    \label{fig:rmse_halfcheetah}
  \end{subfigure}\hspace{0.006\textwidth}%
  \begin{subfigure}[t]{0.32\textwidth}
    \centering
    \includegraphics[width=\linewidth]{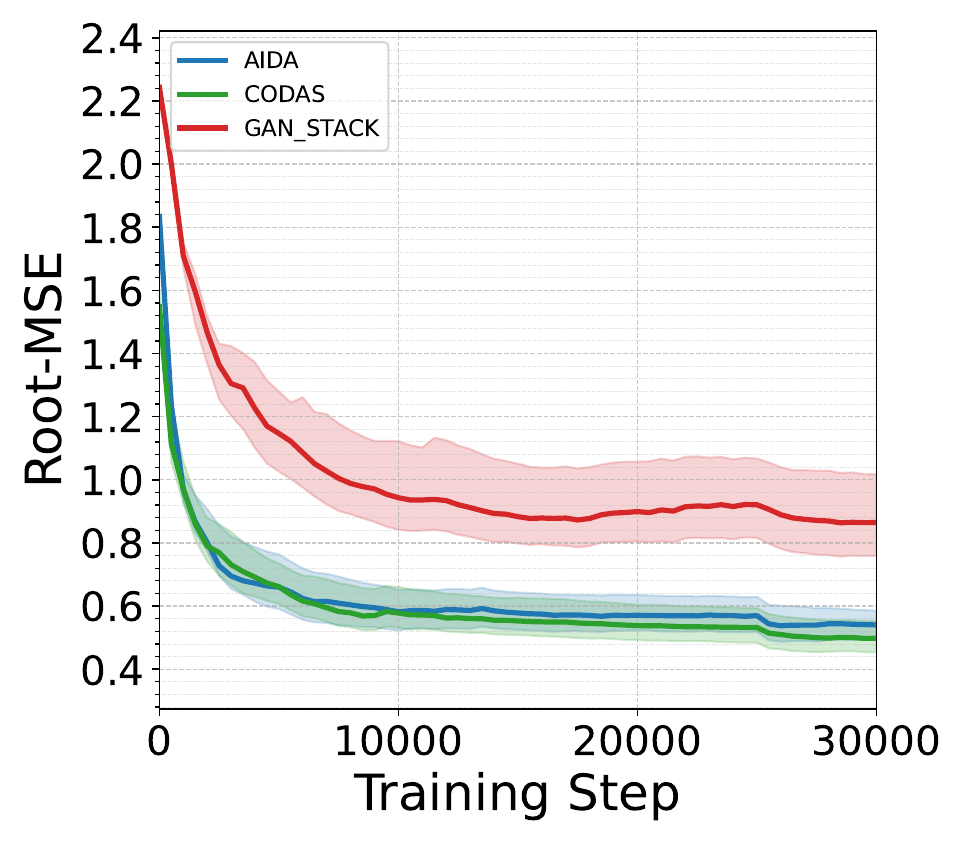}
    \caption{RMSE -- Hopper}
    \label{fig:rmse_hopper}
  \end{subfigure}\hspace{0.006\textwidth}%
  \begin{subfigure}[t]{0.32\textwidth}
    \centering
    \includegraphics[width=\linewidth]{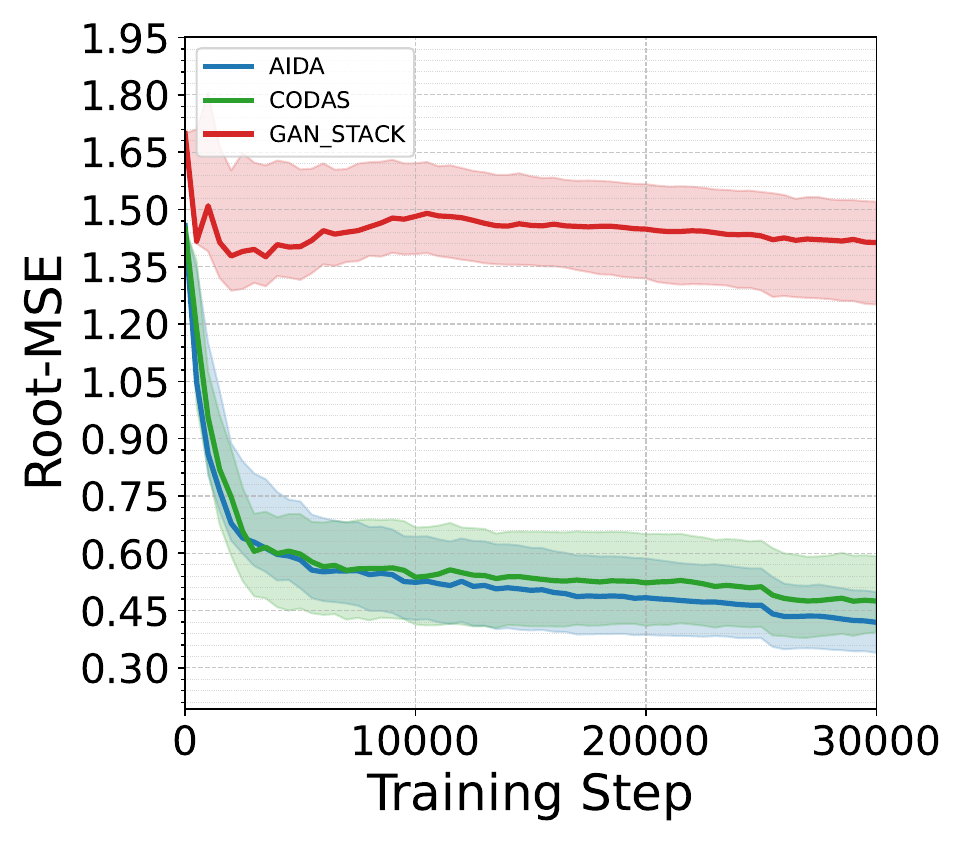}
    \caption{RMSE -- Swimmer}
    \label{fig:rmse_swimmer}
  \end{subfigure}

  \caption{Training curves across three environments. 
  Top row: target-domain return ($\uparrow$) over training iterations. 
  Bottom row: mapping RMSE ($\downarrow$) over training iterations. 
  Solid lines denote the median and shaded regions represent the 
  interquartile range over 3 random seeds.}
  \label{fig:training_curves}
  \vspace{-0.4cm}
\end{figure*}
To assess whether adaptive imagination provides effective additional supervision under scarce target data, we compare AIDA against all baselines across the five MuJoCo tasks and two Gymnasium-Robotics tasks. Note that RMSE directly measures how accurately $q_\phi$ recovers the true proprioceptive state, serving as a quantitative indicator of representation quality.

Table~\ref{tab:main_results} summarizes the results. AIDA achieves the lowest RMSE on every task and the highest return ratio on six out of seven tasks. As shown in Fig.~\ref{fig:training_curves}, AIDA also converges faster than all baselines, with the RMSE curves showing a consistent trend of steeper decrease and stabilization at a lower value, though the difference on Hopper is marginal. 

Comparison with PAD further highlights AIDA's robustness to large domain gaps. Although PAD continuously interacts with the target environment during deployment, it shows lower performance than AIDA across all tasks. This suggests that while PAD's self-supervised adaptation can handle relatively small image-to-image domain gaps, AIDA's adaptation strategy is more effective under the larger state-to-image domain gap considered in our setting.

Notably, on Swimmer and Fetch, AIDA even outperforms Oracle (image), which has access to unlimited online interaction. This is because AIDA benefits from a policy stably trained on low-dimensional states, whereas Oracle (image) must learn both representations and a policy jointly from images. Oracle (image) also achieves relatively low return ratios on Swimmer ($0.304$) and Walker2d ($0.196$), suggesting that these tasks are inherently challenging for extracting meaningful state information from images alone. On Walker2d, all DA methods yield lower return ratios than BC, despite AIDA achieving the best RMSE. We attribute this to the balance-critical nature of Walker2d: even small mapping errors can cause the agent to fall, amplifying the gap between reconstruction accuracy and downstream performance.

Overall, AIDA's advantage stems from two factors. The adaptive imagination mechanism augments the scarce target data with reliable synthetic transitions, and the self-consistency loss leverages these transitions to learn semantically meaningful representations. Together, they provide dense supervision that existing methods, which rely solely on the limited real target data, cannot achieve.
\vskip -0.1cm

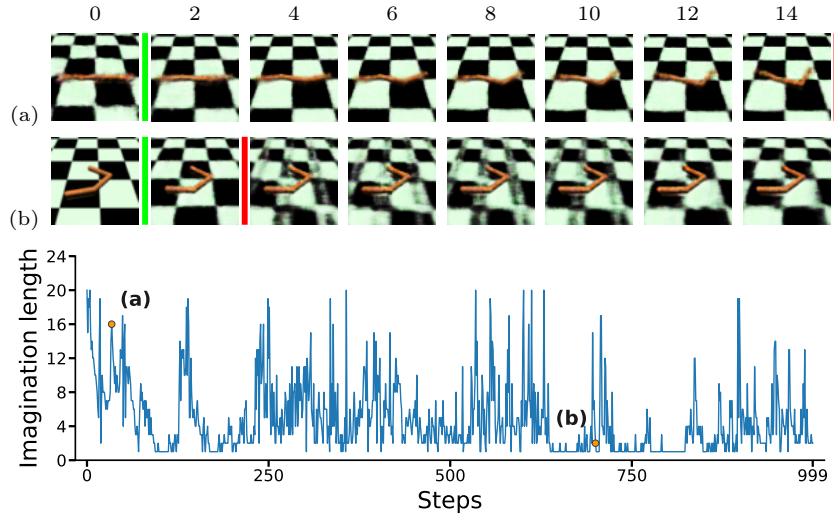
\begin{figure*}[t]
\centering

\newcommand{\iw}{0.095\textwidth}
\newcommand{\im}[2]{\includegraphics[width=\iw]{figure/#1/#2.png}}
\newcommand{\ctximg}[2]{%
  \begin{tikzpicture}[baseline=(img.south)]
    \node[inner sep=0pt] (img) {\im{#1}{#2}};
    \useasboundingbox (img.south west) rectangle (img.north east);
    \draw[green, line width=2pt] ([xshift=\tabcolsep]img.north east) -- ([xshift=\tabcolsep]img.south east);
  \end{tikzpicture}}
\newcommand{\cutimg}[2]{%
  \begin{tikzpicture}[baseline=(img.south)]
    \node[inner sep=0pt] (img) {\im{#1}{#2}};
    \useasboundingbox (img.south west) rectangle (img.north east);
    \draw[red, line width=2pt] ([xshift=\tabcolsep]img.north east) -- ([xshift=\tabcolsep]img.south east);
  \end{tikzpicture}}

\setlength{\tabcolsep}{2pt}

\begin{minipage}{0.90\textwidth}
\centering
\scriptsize

\resizebox{\linewidth}{!}{%
\begin{tabular}{@{}r@{\hspace{4pt}} c c c c c c c c @{}}
& 0 & 2 & 4 & 6 & 8 & 10 & 12 & 14 \\[2pt]
(a)
& \ctximg{6_cut16}{0} & \im{6_cut16}{2}
& \im{6_cut16}{4} & \im{6_cut16}{6}
& \im{6_cut16}{8} & \im{6_cut16}{10}
& \im{6_cut16}{12} & \cutimg{6_cut16}{14} \\[3pt]
(b)
& \ctximg{3_cut3}{0} & \cutimg{3_cut3}{2}
& \im{3_cut3}{4} & \im{3_cut3}{6}
& \im{3_cut3}{8} & \im{3_cut3}{10}
& \im{3_cut3}{12} & \im{3_cut3}{14} \\
\end{tabular}
}

\vspace{2pt}

\includegraphics[width=\linewidth]{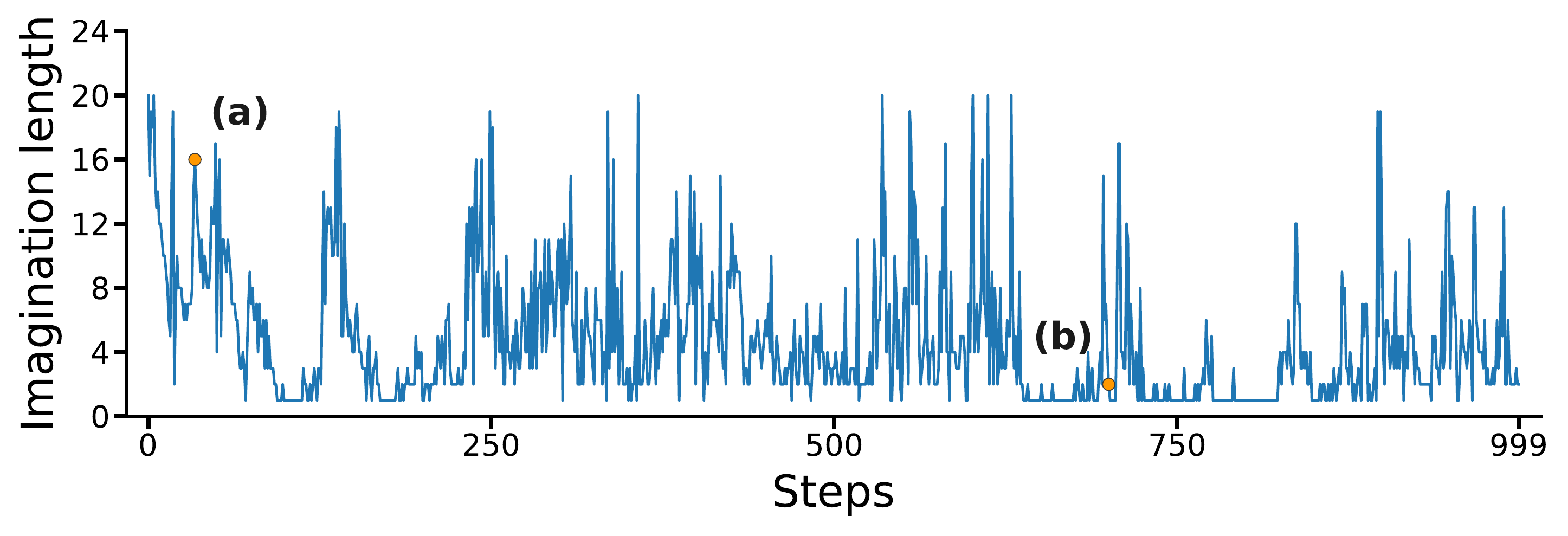}

\end{minipage}

\vskip -0.2cm
\caption{Top: imagined observations decoded from adaptive imagination rollouts on the Swimmer task (shown at every 2 steps).
The {\color{green}green line} marks the point at which imagination begins.
The {\color{red}red line} indicates the truncation point determined by $D_\psi$.
Bottom: adaptive imagination horizon at each step within a single episode, illustrating how the truncation length varies depending on the state encountered at each step.
The two highlighted rollouts, \textbf{(a)} and \textbf{(b)}, correspond to the marked points in the bottom plot (orange dots) and show the decoded imagination sequences starting from those specific steps.}
\vskip -0.4cm
\label{fig:drift_analysis}
\end{figure*}
\subsubsection{(Q2) Does the three-way discriminator reliably detect unreliable drifted transitions?}
\label{drift}

The adaptive truncation mechanism relies on the three-way discriminator $D_\psi$ to detect when imagined transitions drift away from the target-aligned manifold. To verify whether $D_\psi$ makes reasonable truncation decisions, we visualize imagined observations decoded from rollouts across two initial states, and additionally plot the adaptive imagination horizon at each step within a single episode alongside the corresponding observations at selected steps.

As shown in the top of~\cref{fig:drift_analysis}, the decoded observations before the truncation point remain visually coherent and physically plausible, whereas those after the truncation point exhibit noticeable degradation such as distorted body configurations and blurred artifacts. This clear quality gap confirms that the discriminator reliably identifies the boundary between reliable and drifted imagination. Notably, the truncation length varies substantially across states. The bottom of~\cref{fig:drift_analysis} visualizes this per-state variability by plotting the adaptive imagination horizon across steps within a single episode, with the marked points corresponding to the two example rollouts shown above. The horizon fluctuates significantly depending on the encountered state, confirming that imagination reliability is highly state-dependent and motivating adaptive, per-transition truncation rather than a single globally fixed horizon.

\begin{wrapfigure}{r}{0.32\linewidth}
\centering
\includegraphics[width=\linewidth]{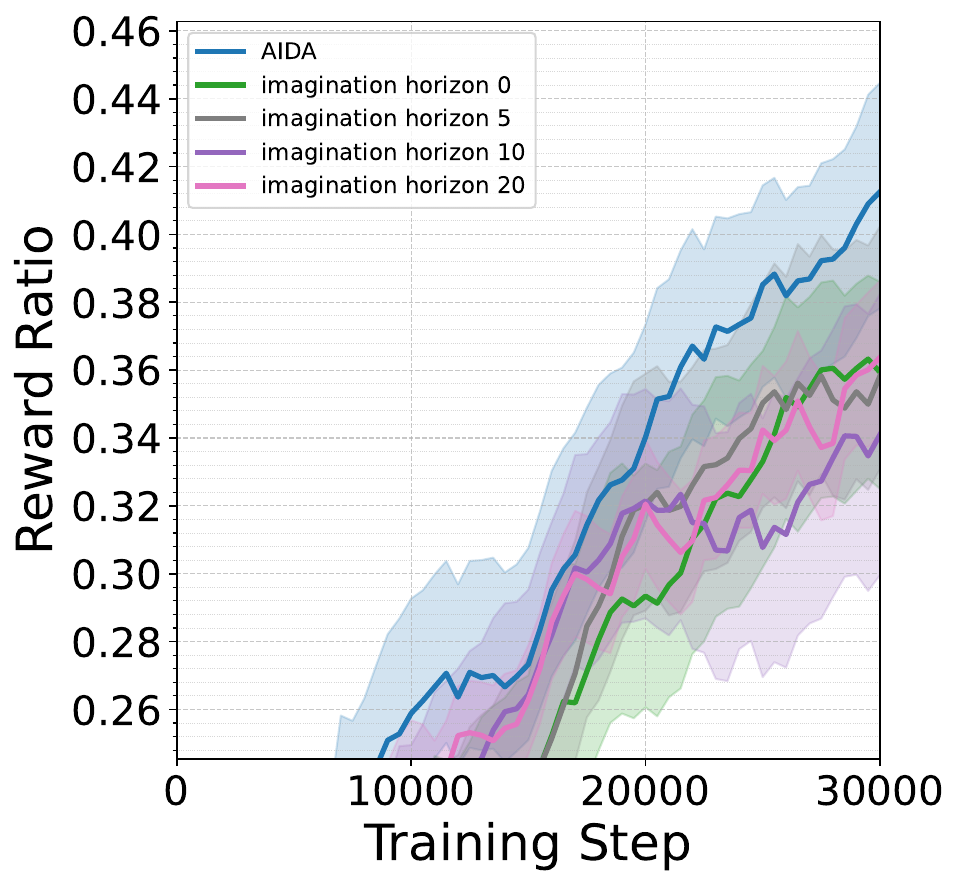}
\caption{Adaptive vs.\ fixed-$K$ training curves on Hopper. 
Return ($\uparrow$). Mean over 3 seeds.}
\label{fig:ablation_curve}
\vspace{-20pt}
\end{wrapfigure}

\subsubsection{(Q3) Does the adaptive imagination outperform fixed-horizon imagination, where unreliable imagined transitions are inevitably included?}
\label{fixed_comparison}

The qualitative analysis in \cref{fig:drift_analysis} revealed that imagination reliability varies significantly across individual states and that the discriminator can detect this drift. We now examine whether including these unreliable transitions actually harms performance, by comparing the adaptive horizon against fixed horizons $K \in \{0, 5, 10, 20\}$.

As shown in Fig.~\ref{fig:ablation_curve}, AIDA achieves the highest return among all configurations on Hopper. Fixed horizons---regardless of length---yield performance comparable to or even worse than no imagination at all, confirming that indiscriminately including unreliable transitions harms rather than helps adaptation. In contrast, the adaptive scheme selectively retains only high-confidence transitions, providing clean supervision that enables meaningful representation learning beyond what scarce target data alone can offer.

\begin{figure}[t]
\centering
\begingroup
\setlength{\tabcolsep}{1pt}
\newcommand{\reconimg}[1]{\includegraphics[width=0.12\linewidth]{#1}}
\begin{tabular}{@{}l@{\hspace{1mm}}cccccc@{}}
\textbf{GT} &
\reconimg{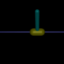} &
\reconimg{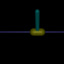} &
\reconimg{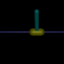} &
\reconimg{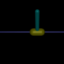} &
\reconimg{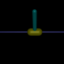} &
\reconimg{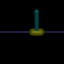} \\
\textbf{AIDA} &
\reconimg{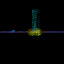} &
\reconimg{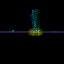} &
\reconimg{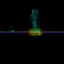} &
\reconimg{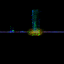} &
\reconimg{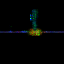} &
\reconimg{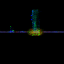} \\
\textbf{CODAS} &
\reconimg{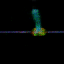} &
\reconimg{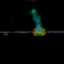} &
\reconimg{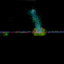} &
\reconimg{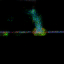} &
\reconimg{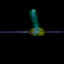} &
\reconimg{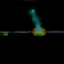} \\
& (a) & (b) & (c) & (d) & (e) & (f) \\
\end{tabular}
\endgroup
\caption{Qualitative comparison of imagined observations on the InvertedPendulum task. Each column shows the ground-truth observation (GT) and those decoded by AIDA (with $\mathcal{L}_{sc}$) and CODAS (without $\mathcal{L}_{sc}$) for the same state.}
\label{fig:recon}
\vskip -0.2cm
\end{figure}

\subsubsection{(Q4) Does the self-consistency loss encourage the encoder to learn state representations that faithfully capture the true physical configuration?}
\label{qualitative_analysis}

To examine whether the self-consistency loss encourages semantically meaningful state representations, we compare the quality of imagined observations from AIDA and CODAS. Since the observation model $p_\theta$ can only render realistic images from states that accurately reflect the true physical configuration, the quality of decoded images directly indicates how well the learned state representations capture the actual physical state. We select InvertedPendulum for this analysis because its reduced trajectory length and image resolution make the difference between methods more pronounced.

As shown in~\cref{fig:recon}, the challenging setting leads to lower reconstruction quality overall, which makes the outputs of AIDA and CODAS appear visually similar. A closer comparison, however, reveals a clear difference. AIDA closely matches the ground-truth pendulum angle and position across all cases, whereas CODAS consistently depicts angles that deviate from the true configuration. This suggests that without the self-consistency loss, the learned state representations may not fully capture the precise physical configuration even when the decoded images appear plausible. The self-consistency loss helps $q_\phi$ learn semantically meaningful state representations that more faithfully reflect the true pendulum orientation.
\vskip -0.3cm
\section{Conclusions}
\label{sec:conclusion}

We presented AIDA, a domain adaptation framework for visual reinforcement learning that leverages adaptive imagination to overcome the challenge of limited target-domain data. AIDA augments scarce target data with reliable and semantically meaningful imagination rollouts. To ensure the reliability of these transitions, a three-way discriminator adaptively truncates drifted rollouts, and a self-consistency loss is trained on the resulting reliable transitions to learn semantically meaningful representations. Experiments on five MuJoCo tasks and two Gymnasium-Robotics tasks demonstrate that AIDA consistently outperforms existing baselines under severely limited target data, the discriminator reliably detects imagination drift, and adaptive truncation outperforms all fixed-horizon alternatives. Despite these results under a challenging cross-modality setting with scarce target data, our work has two limitations. First, the adapted policy still falls short of full source-expert performance, particularly on balance-critical tasks. Incorporating online adaptation algorithm could help close this gap. Second, AIDA assumes shared transition dynamics across domains, which may not hold in practice. Extending AIDA to handle dynamics mismatch is a promising direction for future work.

\section*{Acknowledgements}
This work was supported by the Technology Innovation Program (or Industrial Strategic Technology Development Program-Technology Innovation Program) (RS-2025-25449157, Development of Commercialization Technologies for End-to-End Autonomous Driving Products) funded by the Ministry of Trade, Industry and Resources (MOTIR, Korea), the National Research Foundation of Korea (NRF) grant funded by the Korea government (MSIT) (RS-2026-25497111), the Institute of Information \& Communications Technology Planning \& Evaluation (IITP) under the Artificial Intelligence Convergence Innovation Human Resources Development (IITP-2026-RS-2023-00255968) grant funded by the Korea government (MSIT), and the Ajou University research fund.


%
%
\putbib[main]
\end{bibunit}
\clearpage
\appendix
\begin{bibunit}
\title{Domain Adaptation with Adaptive Imagination for Visual Reinforcement Learning under \\ Limited Target Data: Supplementary Material} 

\titlerunning{AIDA}

\author{Hyunwoo Park\inst{1}\orcidlink{0009-0005-4825-9307} \and
Sang-Hyun Lee\inst{2,3}\orcidlink{0009-0007-3360-4758}}

\authorrunning{H.~Park and S.~Lee}

\institute{STRADVISION, Seoul, Republic of Korea \\
\email{hyunwoo.park@stradvision.com} \and
Department of Mobility Engineering, Ajou University, Suwon, Republic of Korea
\email{sanghyunlee@ajou.ac.kr}}

\maketitle

\section{Training Details}
\subsection{Details of Trajectory Alignment Loss}

The trajectory alignment loss trains the image-to-state mapping $q_\phi$ so that inferred states are visually consistent with the original images and distributionally aligned with source state trajectories. This alignment operates at the trajectory level because a single observation alone can map to multiple plausible states. Sequential context resolves this ambiguity, allowing the loss to reject incorrect mappings through temporal dependencies.

\begin{figure}[tb] 
\centering 
\includegraphics[width=0.7\linewidth]{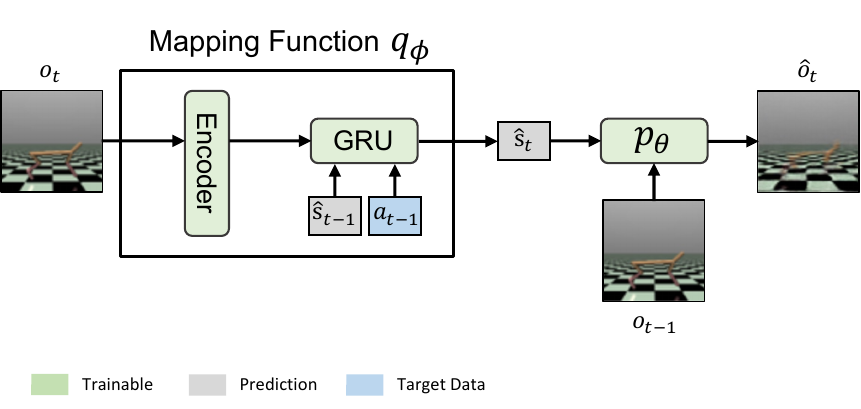} 
\caption{Reconstruction path. The mapping function $q_\phi$ maps the target observation $o_t$ to a inferred state $\hat{s}_t$, and the observation model $p_\theta$ reconstructs the observation $\hat{o}_t$ from $\hat{s}_t$.}
\label{fig:recon_loss} 
\end{figure}

\begin{figure}[tb] 
\centering 
\includegraphics[width=0.90\linewidth]{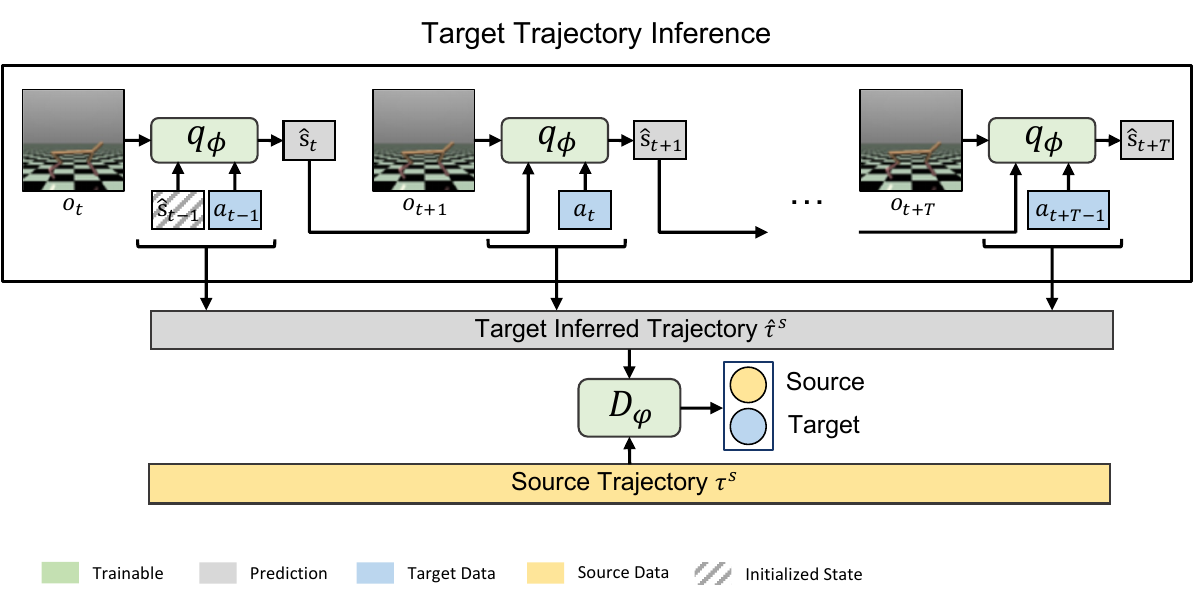} 
\caption{Trajectory alignment path. The mapping function $q_\phi$ sequentially maps target observations to latent states, and the discriminator $D_\eta$ classifies whether the resulting trajectory is from the source or target domain.}
\label{fig:gan_loss} 
\end{figure}

As described in Sec.~3.2 of the main paper, the trajectory alignment loss $\mathcal{L}_{\mathrm{align}}$ consists of two components: a reconstruction loss $\mathcal{L}_{\mathrm{recon}}$ and an adversarial alignment loss $\mathcal{L}_{\mathrm{adv}}$. The reconstruction loss is computed as the mean squared error between the original target observation $o_t$ and the reconstructed observation $\hat{o}_t = p_\theta(\cdot \mid \hat{s}_t, o_{t-1})$:
\begin{equation}
\label{eq:recon}
\mathcal{L}_{\mathrm{recon}} = \frac{1}{T} \sum_{t=1}^{T} \left\| o_t - \hat{o}_t \right\|_2^2.
\end{equation}
This ensures that the inferred state $\hat{s}_t$ retains sufficient information to reconstruct the original target observation. The adversarial loss $\mathcal{L}_{\mathrm{adv}}$ follows the standard GAN loss from Eq.~(1) of the main paper, where the binary discriminator $D_\eta$ is implemented recurrently over transitions to enforce the trajectory-level alignment. For the detailed network architecture, refer to \cref{fig:recon_loss} and \cref{fig:gan_loss}, respectively.

\subsection{Detailed Training Procedure}

The overall training procedure of AIDA is summarized in \cref{alg:training}. 

\paragraph{Stage 1: Source-domain learning (Lines 2--11).} The policy $\pi$ and dynamics model $f_\omega$ are trained in the source domain. At each time step, the agent collects a transition $(s_t, a_t, r_t, s_{t+1})$ and stores it in the replay buffer $\mathcal{B}$ (Lines 5--7). The dynamics model $f_\omega$ is updated via one-step prediction loss (Line 8), and the policy $\pi$ and critic $Q$ are updated via SAC~\cite{haarnoja2018soft} (Line 9). Once converged, both $\pi$ and $f_\omega$ are frozen (Line 12).

\paragraph{Stage 2: Target-domain adaptation (Lines 13--40).}

\paragraph{Trajectory alignment (Lines 14--24).} 
\vspace{-10pt}Given only the pre-collected datasets $\mathcal{D}_{\mathrm{src}}$ and $\mathcal{D}_{\mathrm{tgt}}$, the mapping function $q_\phi$, observation model $p_\theta$, and discriminators $D_\eta$, $D_\psi$ are trained without any environment interaction. At each iteration, a batch of target and source trajectory segments of length $T$ are sampled (Lines 14--15). The mapping function $q_\phi$ sequentially infers states (Lines 16--18).
\begin{algorithm}[H]
\caption{AIDA Training Procedure}
\label{alg:training}
\begin{algorithmic}[1]
\Require Max imagination horizon $K_{\max}$, truncation threshold $\delta$, trajectory length $T$, update frequencies: $I_R$ (reconstruction), $I_D$ (binary discriminator), $I_M$ (mapping)
\State Initialize $f_\omega$, $\pi$, $Q$, $q_{\phi}$, $p_{\theta}$, $D_\eta$, $D_\psi$, replay buffer $\mathcal{B} \gets \varnothing$

\Statex \textbf{Stage 1: Source-domain learning}
\While{not converged}
    \State $s_1 \gets \mathrm{env.reset}()$
    \For{$t=1, \ldots, N$}
        \State $a_t \sim \pi(\cdot \mid s_t)$, \quad $(r_t, s_{t+1}) \gets \mathrm{env.step}(a_t)$
        \State $\mathcal{B} \gets \mathcal{B} \cup \{(s_t, a_t, r_t, s_{t+1})\}$
        \State Sample minibatch from $\mathcal{B}$
        \Comment{Dynamics model and RL agent update}
            \State Update $f_\omega$ via $\mathcal{L}_{\mathrm{wm}} = \| f_\omega(s,a) - s' \|_2^2$
        
        \State Update $\pi$, $Q$ via SAC actor-critic updates
    \EndFor
\EndWhile
\State Freeze $\pi$ and $f_\omega$

\Statex \textbf{Stage 2: Target-domain adaptation (no environment interaction)}
\While{not converged}
    \State Sample $B$ target trajectory segments $\tau^{\mathrm{tgt}} \sim \mathcal{D}_{\mathrm{tgt}}$
    \State Sample $B$ source trajectory segments $\tau^{\mathrm{src}} \sim \mathcal{D}_{\mathrm{src}}$
    \For{$t=1, \ldots, T$} \Comment{Sequential state inference}
        \State $\hat{s}_t \sim q_{\phi}(s_t \mid \hat{s}_{t-1}, a_{t-1}, o_t)$
    \EndFor
    \For{$i=1, \ldots, I_R$} \Comment{Reconstruction updates}
        \State Update $q_{\phi}$ and $p_{\theta}$ via $\mathcal{L}_{\mathrm{recon}}$
    \EndFor
    \For{$i=1, \ldots, I_D$} \Comment{Binary discriminator update}
        \State Update $D_\eta$ via $\mathcal{L}_{\mathrm{adv}}$
    \EndFor
    
    \Statex \hspace{\algorithmicindent} \textcolor{gray}{\textit{// Adaptive imagination with self-consistency}}
    \For{$i=1, \ldots, I_M$} \Comment{Mapping function updates}
        \State Sample rollout start $t_0 \sim \mathrm{Unif}(\{1,\ldots,T\})$
        \State $\tilde{s}_{0} \gets \hat{s}_{t_0}$, \; $\tilde{o}_{0} \gets o_{t_0}$, \; $\mathcal{L}_{\mathrm{sc}} \gets 0$, \; $K^* \gets 0$
        \For{$k=0, \ldots, K_{\max}-1$}
            \State $a_{k} \sim \pi(\cdot \mid \tilde{s}_{k})$, \quad $\tilde{s}_{k+1} \gets f_\omega(\tilde{s}_{k}, a_{k})$ \Comment{$\pi$, $f_\omega$ frozen}
            \If{$D_\psi^{\mathrm{tgt}}(\tilde{s}_{k}, a_{k}, \tilde{s}_{k+1}) < \delta$} \textbf{break} \Comment{Truncate}
            \EndIf
            \State $\tilde{o}_{k+1} \sim p_{\theta}(\cdot \mid \tilde{s}_{k+1}, \tilde{o}_{k})$ \Comment{Decode}
            \State $\bar{s}_{k+1} \sim q_{\phi}(s_{k+1} \mid \tilde{s}_{k}, a_{k}, \tilde{o}_{k+1})$ \Comment{Re-infer}
            \State $\mathcal{L}_{\mathrm{sc}} \mathrel{+}= \|\bar{s}_{k+1}-\tilde{s}_{k+1}\|_2^2$
            \State $K^* \gets k+1$
        \EndFor
        \State Update $q_{\phi}$ and $p_{\theta}$ via $\mathcal{L}_{\mathrm{sc}} / K^*$ \Comment{Self-consistency}
        \State Update $q_\phi$ via $\mathcal{L}_\mathrm{adv}$
        \State Update $D_\psi$ on \{source, target-inferred, imagined\} transitions
    \EndFor
\EndWhile
\end{algorithmic}
\end{algorithm}
\noindent The reconstruction loss is then used to update $q_\phi$ and $p_\theta$ for $I_R$ iterations (Lines 19--21), and the adversarial loss is used to update $D_\eta$ for $I_D$ iterations (Lines 22--24).

\paragraph{Adaptive imagination (Lines 25--40).} The imagination and self-consistency update is repeated $I_M$ times per iteration (Lines 25--39). At each repetition, a rollout start index is sampled uniformly from the trajectory segment (Line 26), and imagination proceeds from the corresponding target-inferred state using the frozen policy and dynamics model (Line 29). At each imagination step, the three-way discriminator evaluates the transition. If the target-likeness score falls below the threshold $\delta$, the rollout is truncated (Lines 30--31). For each reliable imagined step, the observation model decodes an imagined observation and the mapping function re-infers a state, accumulating the self-consistency loss (Lines 32--34). After the rollout, $q_\phi$ and $p_\theta$ are updated via $\mathcal{L}_{\mathrm{sc}} / K^*$ (Line 37). The adversarial loss is used to update $q_\phi$ (Line 38). Finally, the three-way discriminator $D_\psi$ is updated using transitions from all three classes: target-inferred, source, and imagined (Line 39).

\section{Implementation Details}

\paragraph{Pre-processing.}
Source state statistics (mean and standard deviation) are computed from $\mathcal{D}_{\mathrm{src}}$, with a variance floor of $10^{-6}$. The same statistics are used for normalizing both the source states and the inferred states. Target RGB images are obtained via \texttt{env.render()}, resized to $32 \times 32$ for InvertedPendulum and $64 \times 64$ for all other tasks, and scaled to $[0,1]$.

\paragraph{Trajectory processing.}
Each trajectory has a maximum length of 500 steps and is split into rollout chunks of length 100, except for InvertedPendulum, where the trajectory length and rollout chunk length are set to 45 and 15, respectively. Discriminator and reconstruction losses are computed only on valid timesteps within each chunk. The first action and state are zero-initialized at the beginning of each chunk.

\paragraph{Dynamics model.}
The dynamics model predicts the next state in residual form~\cite{he2016deep}. Given a state-action pair $(s, a)$, it outputs a state increment and updates the state as $s' = s + f_\omega(s, a)$. Predicting state differences rather than absolute next states is commonly used in learned dynamics models, and prior work reports that this parameterization can reduce the output range and stabilize training~\cite{xu2025neural}. This property is particularly important in our setting, because AIDA relies on the learned dynamics model to generate imagination rollouts for target-domain adaptation. If one-step prediction is unstable, errors can quickly accumulate during imagination and degrade the reliability of the generated transitions. Therefore, improving one-step prediction stability is critical for producing useful imagination rollouts in practice.
\paragraph{Optimizer and learning rate schedule.}
All modules use the Adam optimizer, except for the discriminators and generator (mapping function), which use RMSprop. The learning rate follows a polynomial decay schedule, linearly decreasing to $50\%$ of the initial value by the end of training.

\paragraph{Training stabilization.}
Several techniques are used to stabilize GAN training:
\begin{itemize}
    \item LayerNorm is applied within the mapping function and the discriminators.
    \item An entropy regularization term $-0.001 \cdot H(\mathrm{logits})$ is added to the adversarial loss.
    \item Gradient norm clipping is applied to discriminator updates. For the generator, both gradient norm clipping and weight-value clipping are used.
    \item Discriminator updates are skipped when it becomes too strong (\i.e., the mean fake probability drops below $0.4$), with a forced update every 20 steps to prevent stalling.
\end{itemize}

\paragraph{Hyperparameters.}
\Cref{tab:hyperparams_s1,tab:hyperparams_s2} lists the hyperparameters used across all experiments. The same values are used for all five tasks unless otherwise noted.

\begin{table}[b]
\centering
\caption{Hyperparameters for AIDA (Stage 1: Source-domain learning).}
\label{tab:hyperparams_s1}
\small
\begin{tabular}{@{}llc@{}}
\toprule
\textbf{Type} & \textbf{Name} & \textbf{Value} \\
\midrule
\multirow{4}{*}{General}
& SAC learning rate (actor / critic) & $1\!\times\!10^{-3}$ / $1\!\times\!10^{-3}$ \\
& Replay buffer size & $1\!\times\!10^{6}$ \\
& Batch size & 256 \\
& Discount factor $\gamma$ & 0.99 \\
\midrule
\multirow{4}{*}{Dynamics Model}
& Hidden layers & {[1024, 1024]} \\
& Activation function & ReLU \\
& Learning rate & $1\!\times\!10^{-3}$ \\
& Minibatch size & 256 \\
\bottomrule
\end{tabular}
\end{table}

\subsection{Environment Details}

We evaluate AIDA on seven continuous control tasks: five tasks from the MuJoCo benchmark suite~\cite{todorov2012mujoco}, implemented via OpenAI Gym, and two reaching tasks from Gymnasium-Robotics, Shadow Dexterous Hand Reach and Fetch Reach. \cref{tab:env_details} summarizes the key properties of each environment. All tasks use continuous action spaces with values in $[-1, 1]$, and the maximum episode length is 1{,}000 steps. As described in the main paper, the source domain provides low-dimensional proprioceptive states (4--63 dimensions depending on the task), while the target domain provides high-dimensional image observations($32 \times 32 \times 3 = 3{,}072$ dimensions for InvertedPendulum and $64 \times 64 \times 3 = 12{,}288$ dimensions for all other tasks). This cross-modality setting offers a practical advantage. The policy can be trained on low-dimensional states, avoiding the instability and computational cost of visual RL. However, it also causes a large dimensionality gap between the two domains. As a result, the adaptation problem becomes considerably harder than the image-to-image setting adopted by most prior work. In particular, the mapping $q_\phi$ must infer precise physical quantities, such as joint angles and velocities, from raw pixels.

\begin{table}[H]
\centering
\caption{Hyperparameters for AIDA (Stage 2: Target-domain adaptation).}
\label{tab:hyperparams_s2}
\small
\begin{tabular}{@{}llc@{}}
\toprule
\textbf{Type} & \textbf{Name} & \textbf{Value} \\
\midrule
\multirow{5}{*}{General}
& $I_D$ & 1 \\
& $I_R$ & 1 \\
& $I_M$ & 5 \\
& Trajectories per iteration & 10 \\
& Trajectory length $T$(InvPend) & 45 \\
& Trajectory length $T$(Except InvPend) & 500 \\ 
\midrule
\multirow{7}{*}{\shortstack[l]{Discriminator\\$D_\eta$}}
& Hidden layers & {[256, 256, 256]} \\
& Activation function & ReLU \\
& Learning rate & $5\!\times\!10^{-5}$ \\
& RNN type & GRU \\
& RNN layers & {[128]} \\
& $H_{\mathrm{reset}}$ & 100 \\
& Adversarial weight $\alpha$ & 2.0 \\
\midrule
\multirow{5}{*}{\shortstack[l]{Three-way\\Discrim.~$D_\psi$}}
& Hidden layers & {[256, 256, 256]} \\
& Activation function & ReLU \\
& Learning rate & $1\!\times\!10^{-3}$ \\
& Max imagination horizon $K_{\max}$ & 20 \\
& Self-consistency weight $\lambda$ & 1.0 \\
\midrule
\multirow{8}{*}{\shortstack[l]{Mapping\\Function $q_\phi$}}
& \textbf{Encoder} & \\
& \quad Conv layers & \shortstack[c]{Conv(4,4,32), Conv(4,4,64),\\Conv(4,4,128), Conv(4,4,256)} \\
& \quad MLP hidden layers & {[256, 256, 256, 256]} \\
& \quad Activation function & LeakyReLU \\
& \textbf{RNN Cell} & \\
& \quad RNN type & GRU \\
& \quad RNN layers & {[128, 128]} \\
\midrule
\multirow{4}{*}{\shortstack[l]{Observation\\Model $p_\theta$}}
& \textbf{Visual Decoder} & \\
& \quad Hidden layers & {[1024, 2048]} \\
& \quad Deconv layers & \shortstack[c]{Deconv(5,5,128), Deconv(5,5,64),\\Deconv(6,6,32), Deconv(6,6,3)} \\
\midrule
\multirow{2}{*}{Train}
& Learning rate ($q_\phi$, $p_\theta$) & $1\!\times\!10^{-4}$ \\
& Training iterations & 30{,}000 \\
\bottomrule
\end{tabular}
\end{table}
\vskip -0.2cm
\begin{table}[H]
\centering
\caption{Summary of environments used in our experiments.}
\label{tab:env_details}
\begin{tabular}{@{}lccl@{}}
\toprule
Environment & State dim. & Action dim. & Description \\
\midrule
HalfCheetah & 17 & 6 & 2D cheetah running forward \\
Hopper & 11 & 3 & 2D one-legged hopping \\
Swimmer & 8 & 2 & 2D multi-link swimming \\
Walker2d & 17 & 6 & 2D bipedal walking \\
InvertedPendulum & 4 & 1 & Balancing an inverted pole \\
Shadow-Reach & 63 & 20 & Dexterous hand fingertip reaching \\
Fetch-Reach & 10 & 4 & Robotic arm end-effector reaching \\
\bottomrule
\end{tabular}
\end{table}


We briefly describe each task below to illustrate the diversity of state representations that $q_\phi$ must infer from images. HalfCheetah is a 2D robot consisting of 9 body parts and 8 joints. The state includes joint angles and velocities (17 dimensions), and the agent applies torques to 6 joints to run forward as fast as possible. Hopper is a 2D one-legged figure with 4 body parts. The 11-dimensional state includes positions and velocities of each body part, and the agent controls 3 hinge joints to hop forward. Swimmer is a 2D multi-link robot in a viscous fluid. The state is 8-dimensional, consisting of joint angles and velocities, and the agent applies torques to 2 joints to swim forward. Walker2d is a 2D bipedal robot with a structure similar to Hopper but with two legs. The 17-dimensional state includes joint angles and velocities, and the agent controls 6 joints to walk forward while maintaining balance. InvertedPendulum is the simplest task, where the agent must balance an inverted pole on a cart. The state is 4-dimensional (cart position, cart velocity, pole angle, pole angular velocity), and the agent applies a single continuous force to the cart. Fetch-Reach requires a robotic manipulator to move its gripper to a target position. The state is 10-dimensional, and the agent controls 4-dimensional gripper displacements. Shadow Dexterous Hand-Reach requires a multi-fingered robotic hand to move its fingertips to target positions. The state is 63-dimensional, and the agent controls 20 actuated degrees of freedom of the hand.

\begin{figure}[tb]
\centering
\captionsetup[subfigure]{labelformat=empty}

\begin{subfigure}[t]{0.23\linewidth}
    \centering
    \includegraphics[width=\linewidth]{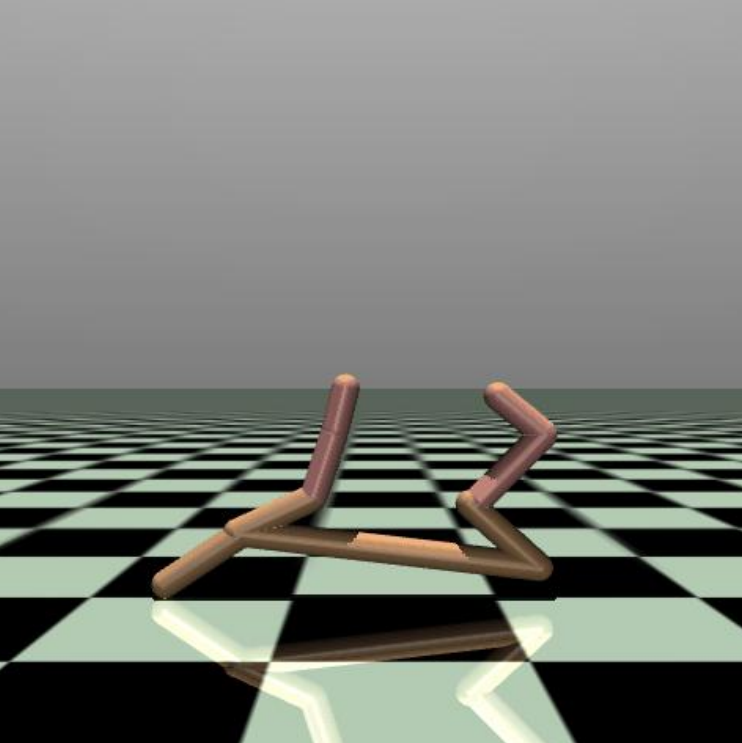}
    \caption{HalfCheetah}
\end{subfigure}
\hfill
\begin{subfigure}[t]{0.23\linewidth}
    \centering
    \includegraphics[width=\linewidth]{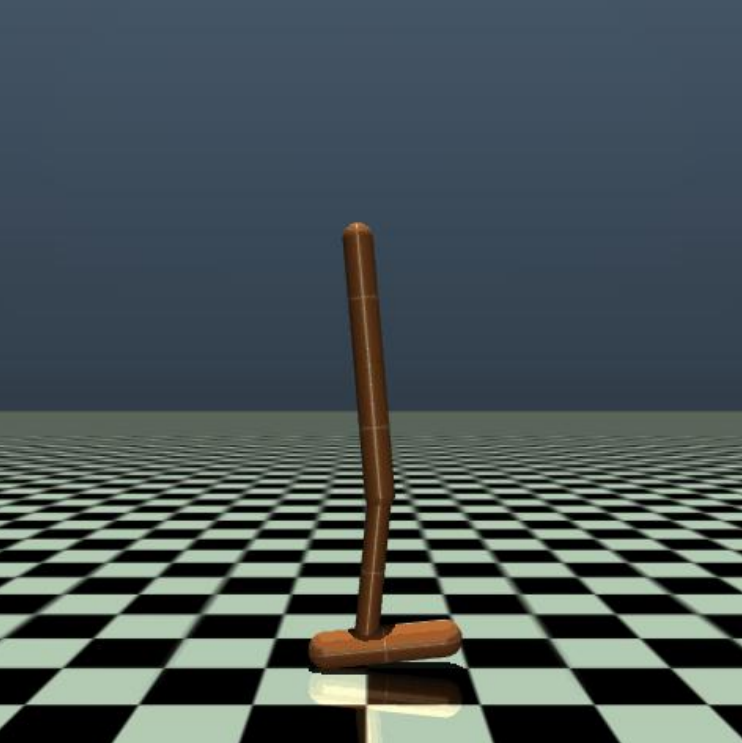}
    \caption{Hopper}
\end{subfigure}
\hfill
\begin{subfigure}[t]{0.23\linewidth}
    \centering
    \includegraphics[width=\linewidth]{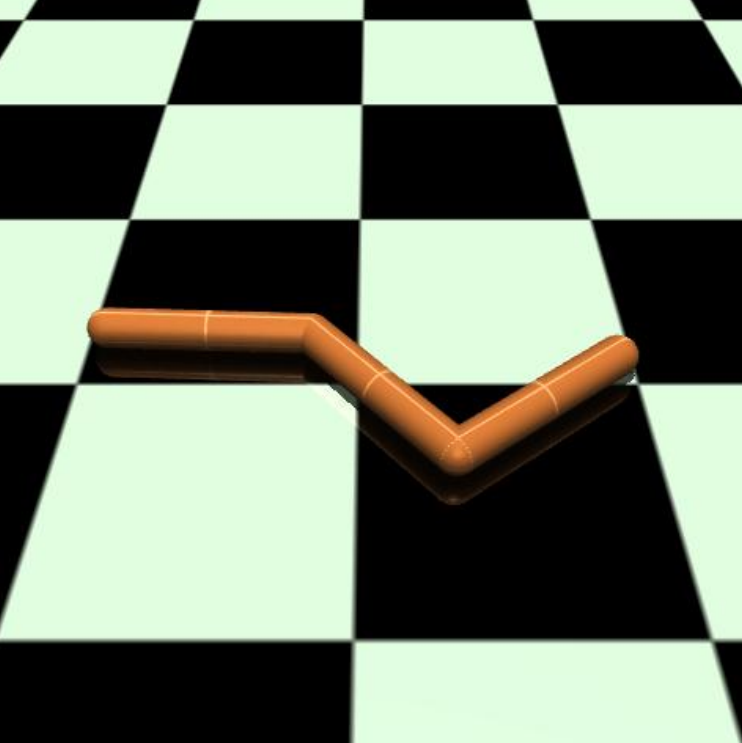}
    \caption{Swimmer}
\end{subfigure}
\hfill
\begin{subfigure}[t]{0.23\linewidth}
    \centering
    \includegraphics[width=\linewidth]{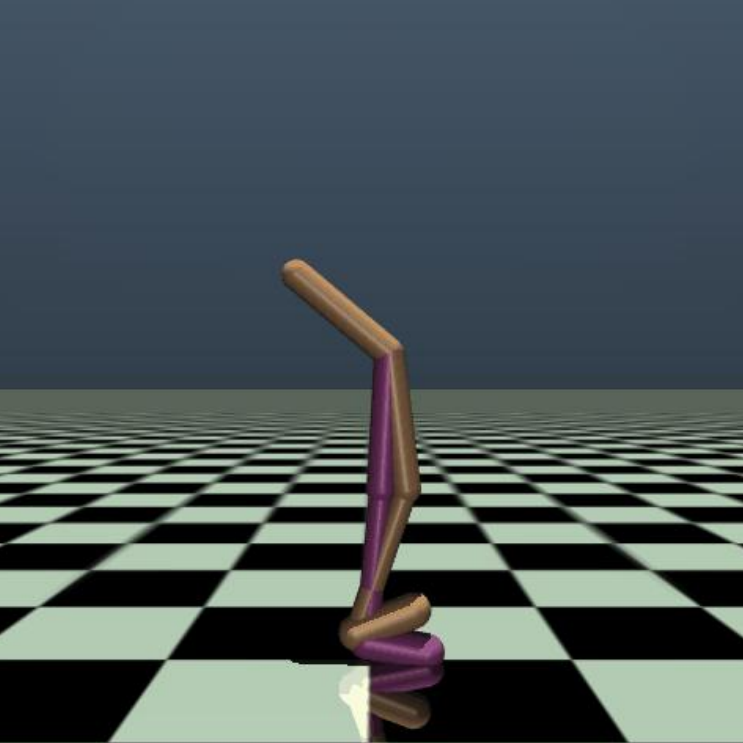}
    \caption{Walker2d}
\end{subfigure}

\vspace{0.5em}

\begin{subfigure}[t]{0.23\linewidth}
    \centering
    \includegraphics[width=\linewidth]{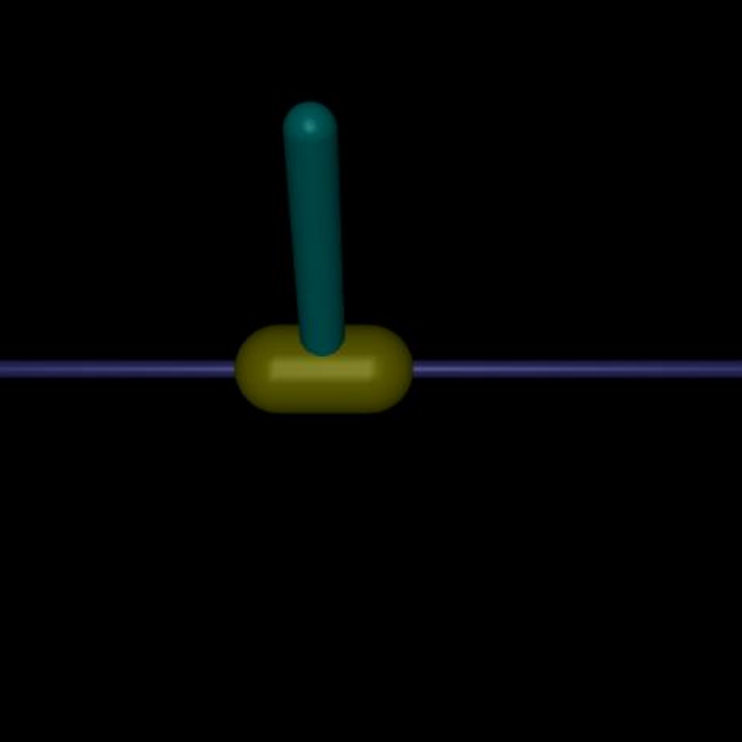}
    \caption{InvertedPendulum}
\end{subfigure}
\hspace{0.01\linewidth}
\begin{subfigure}[t]{0.23\linewidth}
    \centering
    \includegraphics[width=\linewidth]{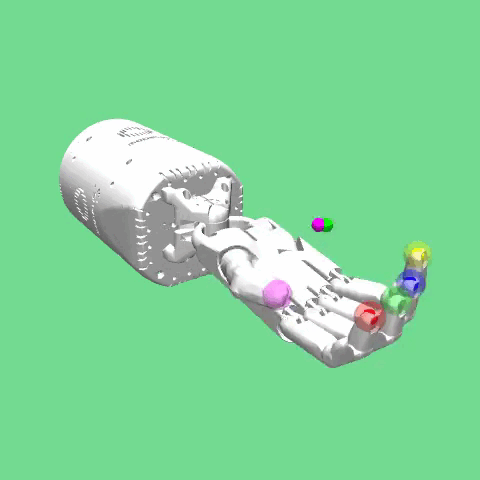}
    \caption{Shadow}
\end{subfigure}
\hspace{0.01\linewidth}
\begin{subfigure}[t]{0.23\linewidth}
    \centering
    \includegraphics[width=\linewidth]{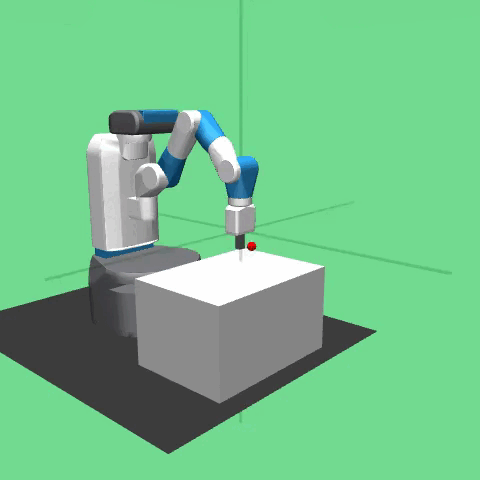}
    \caption{Fetch}
\end{subfigure}

\caption{Example rendered images from each task used in our experiments.}
\label{fig:env_examples}
\end{figure}

The tasks span a range of difficulty in terms of state dimensionality (4 to 63), action dimensionality (1 to 20), and control complexity (static balancing to dexterous hand control), providing a comprehensive testbed for evaluating domain adaptation under cross-modality and scarce data conditions. Illustrations of each task are shown in \cref{fig:env_examples}.

\vskip -0.2cm

\subsection{Data Collection}

To reflect a realistic deployment scenario where real-world data collection is costly and time-consuming, we adopt a limited target data setting. Accordingly, we pre-collect 50 trajectories for all tasks (each with a maximum rollout length of 1{,}000 steps). The expert policy for data acquisition is trained via SAC~\cite{haarnoja2018soft} with low-dimensional state inputs until convergence. For each task, we use the expert policy to collect trajectories in the simulator and render images. The images are rendered using the default camera configuration via \texttt{render\_mode="rgb\_array"} and resized to $32 \times 32 \times 3$ for InvertedPendulum and $64 \times 64 \times 3$ for all other tasks. The collected state-action trajectories are regarded as the source dataset $\mathcal{D}_{\mathrm{src}}$, and the rendered image-action trajectories are regarded as the target dataset $\mathcal{D}_{\mathrm{tgt}}$. Examples of rendered images for each task are shown in \cref{fig:env_examples}.

Although $\mathcal{D}_{\mathrm{src}}$ can be much larger than $\mathcal{D}_{\mathrm{tgt}}$, we deliberately subsample it to match the size of $\mathcal{D}_{\mathrm{tgt}}$ at each training iteration. This follows the standard GAN training practice where the batch sizes of real and generated samples are kept equal~\cite{guo2020positive}. Without this balancing, the discriminator quickly overfits to the majority class and provides uninformative gradients to the generator.

\section{Additional Experiment Results}


\begin{figure}[tb] 
\centering 
\includegraphics[width=0.7 \linewidth]{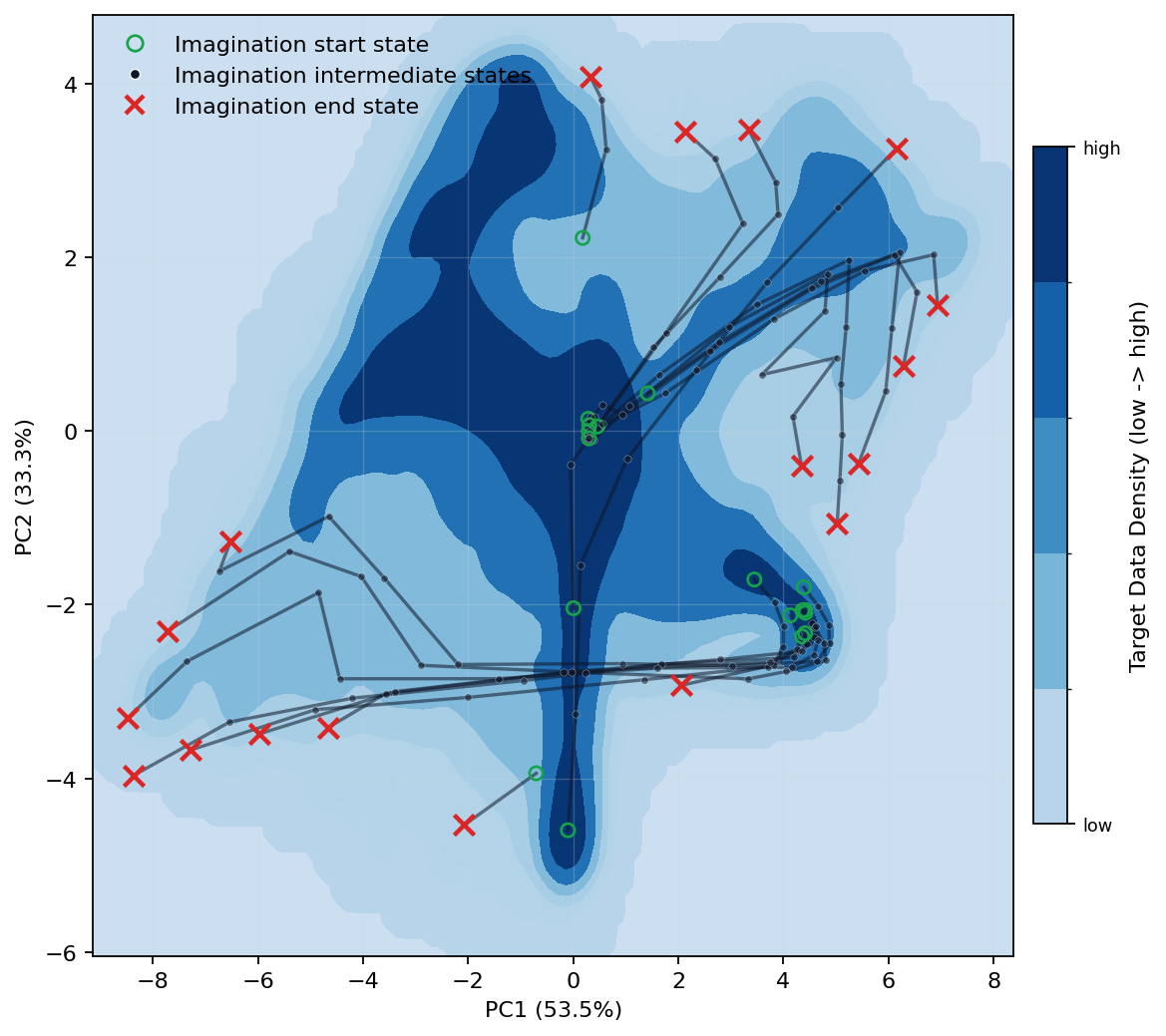}
\caption{PCA visualization of imagination rollouts on the Swimmer task. The background contour shows the kernel density estimate of the target data distribution. Green circles denote the start of each rollout, black dots are intermediate imagined states, and red crosses mark the truncation points determined by $D_\psi$.}
\label{fig:three-way_discriminator_distribution_drift} 
\end{figure}
\subsection{Additional Analysis of Three-way Discriminator}

To verify that the three-way discriminator truncates rollouts that drift away from the target data distribution, we visualize imagination rollouts in state space using PCA. \cref{fig:three-way_discriminator_distribution_drift} shows the first two principal components of the state space, with the background contour representing the kernel density estimate of the target data distribution. Each rollout is depicted as a trajectory from a start state (green circle) through intermediate imagined states (black dots) to the truncation point (red cross) determined by $D_\psi$.

As shown in the figure, most rollouts begin within or near the high-density regions of the target distribution, and the three-way discriminator permits imagination to continue as long as the trajectory remains in these regions. Once the imagined states drift into low-density regions, the three-way discriminator triggers truncation. This is evident from the red crosses, which consistently appear in low-density areas of the target distribution. This confirms that $D_\psi$ effectively detects when imagination rollouts leave the target data distribution and truncates them accordingly, preventing unreliable transitions from being used for self-consistency training.

\subsection{Extended Visualization of Imagination Drift}
\begin{figure}[H]
\centering
\newcommand{\iw}{0.095\textwidth}
\newcommand{\im}[2]{\includegraphics[width=\iw]{figure/#1/#2.png}}
\newcommand{\ctximg}[2]{%
  \begin{tikzpicture}[baseline=(img.south)]
    \node[inner sep=0pt] (img) {\im{#1}{#2}};
    \useasboundingbox (img.south west) rectangle (img.north east);
    \draw[green, line width=2pt] ([xshift=\tabcolsep]img.north east) -- ([xshift=\tabcolsep]img.south east);
  \end{tikzpicture}}
\newcommand{\cutimg}[2]{%
  \begin{tikzpicture}[baseline=(img.south)]
    \node[inner sep=0pt] (img) {\im{#1}{#2}};
    \useasboundingbox (img.south west) rectangle (img.north east);
    \draw[red, line width=2pt] ([xshift=\tabcolsep]img.north east) -- ([xshift=\tabcolsep]img.south east);
  \end{tikzpicture}}
\setlength{\tabcolsep}{2pt}
\scriptsize
\begin{tabular}{@{}r@{\hspace{4pt}} c c c c c c c c @{}}
& 0 & 2 & 4 & 6 & 8 & 10 & 12 & 14 \\[2pt]
(a)
& \ctximg{6_cut16}{0} & \im{6_cut16}{2}
& \im{6_cut16}{4} & \im{6_cut16}{6}
& \im{6_cut16}{8} & \im{6_cut16}{10}
& \im{6_cut16}{12} & \cutimg{6_cut16}{14} \\[3pt]
(b)
& \ctximg{1_cut10}{0} & \im{1_cut10}{2}
& \im{1_cut10}{4} & \im{1_cut10}{6}
& \im{1_cut10}{8} & \cutimg{1_cut10}{10}
& \im{1_cut10}{12} & \im{1_cut10}{14} \\[3pt]
(c)
& \ctximg{2_cut8}{0} & \im{2_cut8}{2}
& \im{2_cut8}{4} & \im{2_cut8}{6}
& \cutimg{2_cut8}{8} & \im{2_cut8}{10}
& \im{2_cut8}{12} & \im{2_cut8}{14} \\[3pt]
(d)
& \ctximg{3_cut3}{0} & \cutimg{3_cut3}{2}
& \im{3_cut3}{4} & \im{3_cut3}{6}
& \im{3_cut3}{8} & \im{3_cut3}{10}
& \im{3_cut3}{12} & \im{3_cut3}{14} \\[3pt]
(e)
& \ctximg{5_cut2}{0} & \cutimg{5_cut2}{2}
& \im{5_cut2}{4} & \im{5_cut2}{6}
& \im{5_cut2}{8} & \im{5_cut2}{10}
& \im{5_cut2}{12} & \im{5_cut2}{14} \\[3pt]
(f)
& \ctximg{4_cut19}{0} & \im{4_cut19}{2}
& \im{4_cut19}{4} & \im{4_cut19}{6}
& \im{4_cut19}{8} & \im{4_cut19}{10}
& \im{4_cut19}{12} & \cutimg{4_cut19}{14} \\
\end{tabular}
\vspace{2pt}
\includegraphics[width=1.0\linewidth]{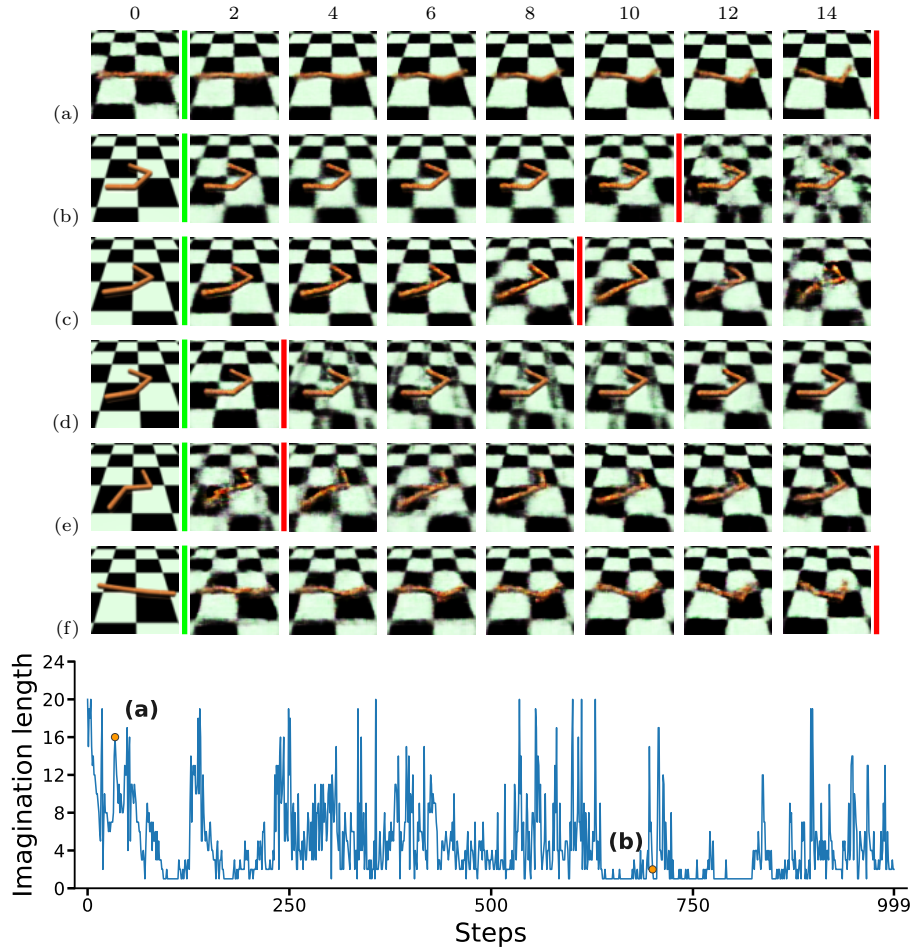}
\vskip -0.2cm
\caption{Extended visualization of imagination drift on the Swimmer task (shown at every 2 steps). The {\color{green}green line} marks the point at which imagination begins. The {\color{red}red line} indicates the truncation point determined by $D_\psi$. The bottom plot shows the adaptive imagination horizon at each step within a single episode. Rollouts (a)--(f) correspond to the marked points (orange dots) in the bottom plot, illustrating how the truncation length varies substantially across states.}
\vskip -0.2cm
\label{fig:drift_analysis_extended}
\end{figure}

In the main paper (Fig.~3), we visualize imagination rollouts for two selected initial states. Here we provide an extended visualization with six initial states to further demonstrate the per-state variability of imagination reliability. As shown in \cref{fig:drift_analysis_extended}, the truncation point varies widely across states: (a) and (f) permit nearly the full rollout, (b) and (c) truncate at moderate horizons, and (d) and (e) truncate almost immediately. In all cases, the decoded observations after the truncation point exhibit visible degradation compared to those before, confirming the discriminator's reliability across diverse states.
\vspace{10pt}

\subsection{Sensitivity to the Amount of Target Data}
We conducted an additional target-data budget analysis on
 HalfCheetah by reducing the number of target trajectories
from the original setting of 50 to 30, 20, and 10. As shown
in \cref{tab:data_budget}, AIDA's return ratio decreases with fewer target trajectories, but its gap over CODAS becomes larger. This suggests that adaptive imagination becomes more effective as target data becomes more limited. 
 \begin{table*}[tb]
\centering
\caption{Return ratio of AIDA and CODAS on HalfCheetah under different target-data budgets.}
\label{tab:data_budget}
\begin{tabular}{@{}lcccc@{}}
\toprule
Target Data Budget & 50 & 30 & 20 & 10 \\
\midrule
AIDA & \textbf{0.810} & \textbf{0.573} & \textbf{0.448} & \textbf{0.175} \\
CODAS & 0.711 & 0.476 & 0.312 & 0.058 \\
\bottomrule
\end{tabular}
\end{table*}


\putbib[supplementary/supplementary]
\end{bibunit}

\end{document}